\documentclass{article}

\usepackage{microtype}
\usepackage{graphicx}
\usepackage{subfigure}
\usepackage{booktabs} 

\usepackage{hyperref}

\usepackage[accepted]{icml2024}

\usepackage{amsmath}
\usepackage{amssymb}
\usepackage{mathtools}
\usepackage{amsthm}

\usepackage[capitalize,noabbrev]{cleveref}

\usepackage{bm}

\usepackage{makecell}
\usepackage{adjustbox}

\theoremstyle{plain}
\newtheorem{theorem}{Theorem}[section]
\newtheorem{proposition}[theorem]{Proposition}
\newtheorem{lemma}[theorem]{Lemma}

\newtheorem{conditions}[theorem]{Conditions}
\theoremstyle{definition}
\newtheorem{definition}[theorem]{Definition}
\theoremstyle{remark}
\newtheorem{remark}[theorem]{Remark}

\theoremstyle{plain} 
\newtheorem{assumption}{Assumption}

\usepackage{arydshln}

\usepackage[textsize=tiny]{todonotes}

\icmltitlerunning{Probabilistic Constrained RL with Formal Interpretability}

\begin{document}

\twocolumn[
\icmltitle{Probabilistic Constrained Reinforcement Learning with Formal Interpretability}






\begin{icmlauthorlist}
\icmlauthor{Yanran Wang}{yyy}
\icmlauthor{Qiuchen Qian}{yyy}
\icmlauthor{David Boyle}{yyy}
\end{icmlauthorlist}

\icmlaffiliation{yyy}{Systems and Algorithms Laboratory, Imperial College London, South Kensington, London}

\icmlcorrespondingauthor{Yanran Wang}{yanran.wang20@imperial.ac.uk}

\icmlkeywords{Machine Learning, ICML}

\vskip 0.3in
]



\printAffiliationsAndNotice{}  

\begin{abstract}
Reinforcement learning can provide effective reasoning for sequential decision-making problems with variable dynamics. Such reasoning in practical implementation, however, poses a persistent challenge in interpreting the reward function and the corresponding optimal policy. Consequently, representing sequential decision-making problems as probabilistic inference can have considerable value, as, in principle, the inference offers diverse and powerful mathematical tools to infer the stochastic dynamics whilst suggesting a probabilistic interpretation of policy optimization. In this study, we propose a novel Adaptive Wasserstein Variational Optimization, namely AWaVO, to tackle these interpretability challenges. Our approach uses formal methods to achieve the interpretability for convergence guarantee, training transparency, and intrinsic decision-interpretation. To demonstrate its practicality, we showcase guaranteed interpretability with a global convergence rate $\Theta(1/\sqrt{T})$ in simulation and in practical quadrotor tasks. In comparison with state-of-the-art benchmarks, including TRPO-IPO, PCPO, and CRPO, we empirically verify that AWaVO offers a reasonable trade-off between high performance and sufficient interpretability.
\end{abstract}

\section{Introduction}
\label{Section_intro}
Sequential decision-making problems can be represented using Reinforcement Learning (RL) or optimal control technologies to efficiently determine optimal policies or control strategies in the presence of uncertainties \cite{levine2018reinforcement}. Nevertheless, such reasoning poses an ongoing challenge to create a convincing interpretation of sequential decision-making and its corresponding optimal policies \cite{devidze2021explicable,levine2022understanding}. This challenge in comprehension poses a significant barrier to real-world RL's implementation in safety-critical domains, such as advanced manufacturing \cite{napoleone2020review}, autonomous navigation \cite{fernandez2023trustworthy} and financial trading \cite{mcnamara2016law}.

\textbf{Key Challenges.} \; The challenges surrounding interpretability in the context of RL can be conceptualized through three distinct phases: \textbf{a.} \textbf{Guarantee of convergence} ensures that an RL framework converges towards an optimal policy, e.g., in an asymptotic manner. \textbf{b.} \textbf{Transparency in training convergence} emphasizes the identification of the underlying processes that an RL algorithm employs to reach convergence during its training. An instance is the convergence rate, where, based on a given number of training iterations, the rate enables the prediction of the expected level of convergence with a certain degree of confidence. \textbf{c.} \textbf{Interpretation of decisions} seeks to explain \textit{why} these specific sequential decisions were made within a given state and environment. Specifically, this interpretation involves clarifying the quantitative impact of latent factors on these sequential decisions. Moreover, due to legal mandates in industries, this facet of interpretation is of even greater significance, particularly in ensuring the trustworthiness of self-driving vehicles \cite{fernandez2021trustworthy,fernandez2023trustworthy}, aerospace engineering \cite{brat2021we,torens2022guidelines}, and high-frequency trading \cite{mcnamara2016law}.

One widely adopted approach to achieving model interpretability involves the use of post-hoc explanation methods. These methods provide retrospective rationales for model predictions, often through the creation of saliency maps or exemplars, as discussed in previous research \cite{lipton2018mythos,kenny2021explaining}. Despite their popularity, these approaches may produce incomplete or inaccurate explanations \cite{slack2020fooling,zhou2022feature}. In response to these limitations, recent studies have shifted their focus towards intrinsic interpretability \cite{rudin2019stop,kenny2022towards}. These methods, however, face challenges in providing a transparent and comprehensive view in the decision-making process. While they present decision explanations that are user-friendly, as demonstrated in \cite{kenny2022towards}, corresponding to \textbf{Key Challenge c}, there is no guaranteed transparency, as highlighted in \textbf{Key Challenges a} and \textbf{b}. Establishing such transparency is crucial and serves as a prerequisite for underpinning user trust and predicting the system's capabilities.

\begin{figure}[t]
    \begin{center}
    \includegraphics[scale=0.405]{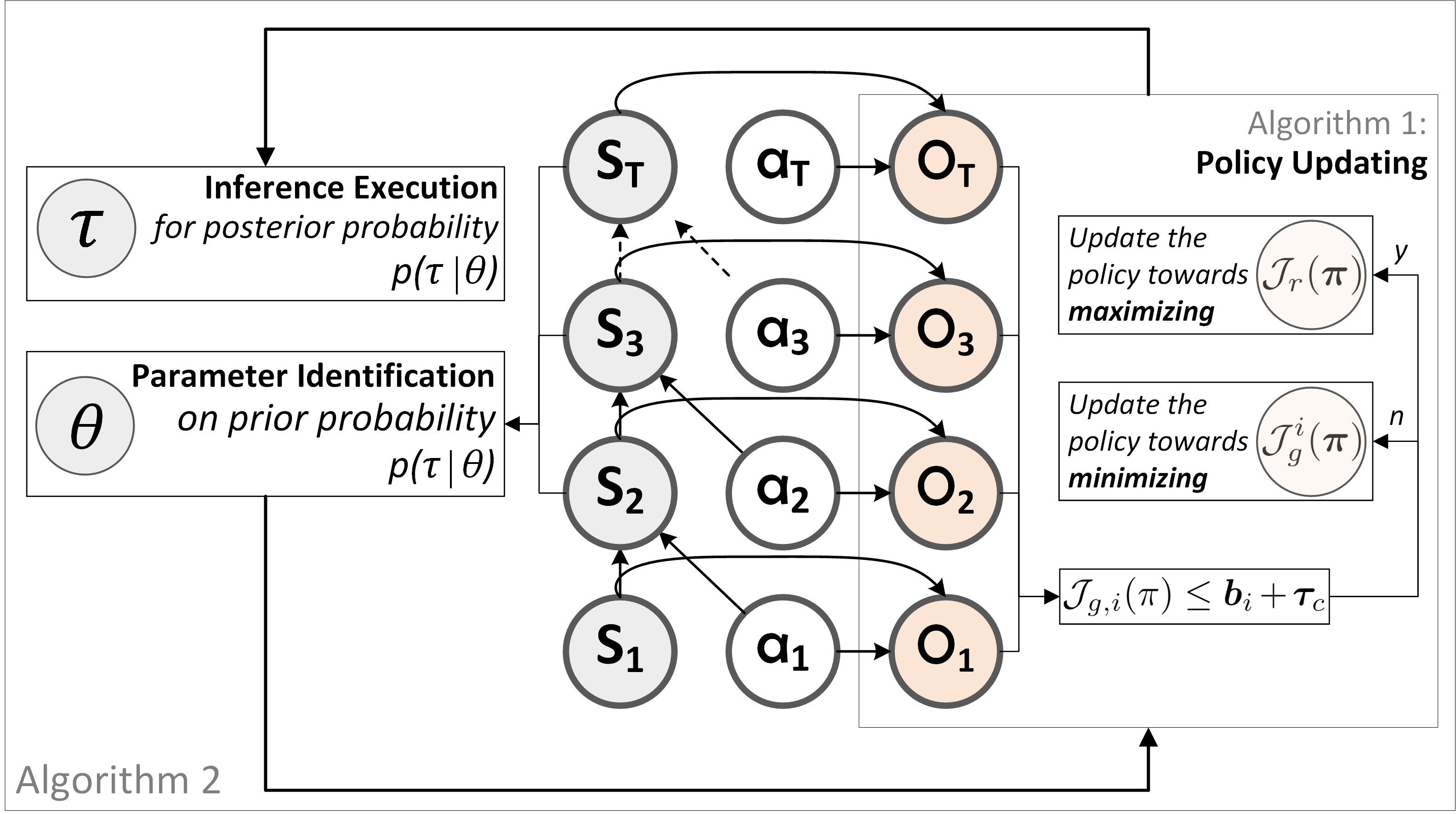}
    \end{center}
    \caption{A new graphical model for constrained RL: refer to Algorithm~{\autoref{AWaVO}} for a comprehensive overview of \textit{(i) Parameter Identification}, \textit{(ii) Policy Updating} and \textit{(iii) Inference Execution}.}
    \label{pgm_saferl}
\end{figure}

To our best knowledge, we present the first intrinsically interpretable constrained RL framework through the lens of probabilistic inference. Specifically, we reframe constrained RL as Wasserstein variational optimization, leveraging an enhanced foundational inference framework known as augmented Probabilistic Graphical Models (PGMs). This is illustrated in \autoref{pgm_saferl}. 
Our proposed \textbf{A}daptive Sliced \textbf{Wa}sserstein \textbf{V}ariational \textbf{O}ptimization (AWaVO), as elaborated in \autoref{Adaptive_slicing_framework}, 
consists of two primary steps:
\begin{itemize}
\item[a.] \textit{Optimality-Rectified Policy Optimization using Distributional Representation (ORPO-DR)}: ORPO is conducted to dynamically adapt to uncertainties (Algorithm~{\autoref{Opt_Rect_Policy_Update}}, Section \ref{Subsec_orpodr}). More importantly, Distributional Representation (DR) provides an entire distribution of the action-value function, contributing to heightened transparency in the convergence process (as outlined in \textbf{\cref{theorem_1}} and \textbf{\cref{theorem_2}}). Consequently, this efficiently tackles a significant portion of the deficiencies outlined in \textbf{Key Challenges a} and \textbf{b};
\item[b.] \textit{Wasserstein Variational Inference (WVI)}: as detailed in Section \ref{Section_WVI}, WVI is subsequently performed to achieve the probabilistic interpretation of decisions, thereby tackling \textbf{Key Challenge c}.
\end{itemize}
Our contributions can be summarized as follows:
\begin{itemize}
\item \textbf{Adaptive Generalized Sliced Wasserstein Distance}, referred to as A-GSWD, incorporates the Sliced Wasserstein Distance (SWD) along with adaptive Radon transforms to handle dynamic uncertainties. Specifically, the proposed A-GSWD adaptively determines the hypersurfaces' slicing directions to enhance the precision of distribution distance computation;
\item \textbf{Adaptive Sliced Wasserstein Variational Optimization}, abbreviated as AWaVO, employs inference to reformulate the problem of sequential decision-making. To tackle all Key Challenges, AWaVO leverages ORPO-DR to enhance the transparency of convergence under dynamic uncertainties. Additionally, WVI is employed to provide probabilistic decision-interpretation;
\item \textbf{Formal methods for interpretation} are employed to demonstrate theoretical comprehension on the metric judgment of A-GSWD, transparency of training convergence, and probabilistic interpretation of decisions. The practical hardware implementation and additional demonstrations are showcased in a video \footnote{\url{https://github.com/Alex-yanranwang/AWaVO}}.
\end{itemize}

\section{Related Work}
\textbf{Reinforcement Learning as Inference.} \; The relationship between sequential decision-making and probabilistic inference has been explored extensively in recent years \cite{levine2018reinforcement,okada2020variational,liu2022constrained}. Despite variations in terminology, the core inference frameworks remain consistent, namely, PGMs \cite{koller2009probabilistic}. While substantial research exists on learning and inference techniques within PGMs \cite{levine2018reinforcement}, the direct connection between RL (or control) and probabilistic inference is not immediately apparent. \cite{welch1995introduction} establishes that control and inference are dual perspectives of the same problem. This connection offers novel insights and enhanced understanding within control problems by leveraging mathematical tools of inference \cite{toussaint2006probabilistic,kappen2012optimal}. Moreover, the study on `RL as inference' represents another prominent trend. Specifically, \cite{levine2018reinforcement} demonstrates that RL is equivalent to probabilistic inference under dynamics. \cite{chua2018deep,okada2020variational} approach dynamics modeling by employing Bayesian inference optimization. Furthermore, \cite{o2020making} revisits the formalization of `RL as inference' and demonstrates that with a slight algorithmic modification, this approximation can perform well even in problems where it initially performs poorly. In this study, we formalize constrained RL as Wasserstein variational optimization to achieve decision-interpretations.

\textbf{Optimal Transport Theory.} \; Forming effective metrics between two probability measures is a fundamental challenge in machine learning and statistics communities. The optimal transport theory, particularly the Wasserstein distance, has garnered significant attention across various domains \cite{solomon2014wasserstein,kolouri2017optimal,schmitz2018wasserstein,wang2023trustworthy} due to its accuracy, robustness, and stable optimization. Nevertheless, due to its computational demand on high-dimensional data, recent advancements emphasize computational efficiency through differentiable optimization \cite{peyre2017computational}. Among these methods, Sinkhorn distance \cite{cuturi2013sinkhorn,altschuler2017near} introduces entropy regularization to smoothen the convex regularization. Another notable approach involves slicing or linear projection \cite{ng2005fourier}, i.e., Sliced Wasserstein Distance (SWD) \cite{bonneel2015sliced}, which leverages the measures' Radon transform for efficient dimensionality reduction. Then, variants of SWD, such as Generalized SWD (GSWD) \cite{kolouri2019generalized}, improve projection efficiency. These advancements contribute to the efficiency in optimal-transport-based metrics. However, they suffer from reduced accuracy as SWD only slices distributions using linear hyperplanes, which may fail to capture the complex structures of data distributions. To overcome the accuracy limitation, Augmented SWD (ASWD) \cite{chen2021augmented} projects onto flexible nonlinear hypersurfaces, enabling the capture of intricate data distribution structures. Building upon the ASWD framework, we introduce an adaptive variant called A-GSWD which leverages the projection onto nonlinear hypersurfaces and combines it with ORPO-DR to achieve adaptivity. This adaptive approach enhances the efficiency and accuracy of Wasserstein distance computation, improving upon the limitations of previous methods.

\section{Problem Formulation and Preliminaries}
\label{Section_seq_decision}
\textbf{Sequential Decision-making as Probabilistic Inference.} \quad A sequential decision-making problem, formalized as a standard RL or control problem, can be seen as an inference problem \cite{levine2018reinforcement}:
\begin{equation}
\begin{aligned}
p&(\tau|\mathcal{O}_{0:T-1}=1)\propto \int  \underbrace{\prod_{t=0}^{T-1}p(\mathcal{O}_{t}=1|\bm{s}_t,\bm{a}_t) }_{:=p(\mathcal{O}|\tau)}  \cdot \\
&\underbrace{ p(\bm{s}_0) \left\{\prod_{t=0}^{T-1}p(\bm{a}_{t}|\bm{s}_t,\theta)p(\bm{s}_{t+1}|\bm{s}_t,\bm{a}_t)\right\}}_{\stackrel{\text{Markov property}}{:=} p(\tau|\theta) } \cdot \underbrace{p(\theta|D)}_{:=p_{D}(\theta)}\mathrm{d}\theta
\label{Bayes_pro}
\end{aligned}
\end{equation}
where $\bm{s}_t$, $\bm{a}_t$, $\tau=\left \{(\bm{s}_t,\bm{a}_t)\right \}_{t=0}^{T-1}$ and $D=\left \{(\bm{s}_t,\bm{a}_t,\bm{s}_{t+1} )\right \}$ are states, actions, a trajectory and observed training dataset. $\mathcal{O}_t=\{\mathcal{O}_{r,t},\mathcal{O}_{g_i,t}\}\in\left\{0,1\right\}$ represents an additional binary variable of the optimality for $(\bm{s}_t,\bm{a}_t)$ in PGM \cite{levine2018reinforcement,okada2020variational}. $\mathcal{O}_{r,t}=1$ and $\mathcal{O}_{g_i,t}=1$ signify that the trajectory $\tau$ is optimized and compliant with the constraints, respectively.

In \autoref{Bayes_pro}, we can deconstruct the various components: \textbf{the probability $p(\bm{a}_{t}|\bm{s}_t,\theta)$} signifies the stationary policy $\pi$ which maps one state $\bm{s}_t$ to one action $\bm{a}_t$, where $\bm{a}_{t}\sim p(\cdot|\bm{s}_t,\theta)=\pi(\cdot|\bm{s}_t)$ at each time step $t$; \textbf{the transition probability $p(\bm{s}_{t+1}|\bm{s}_t,\bm{a}_t)$} represents state transitions (also known as forward-dynamics models), where $\bm{s}_{t+1}\sim p(\cdot|\bm{s}_t,\bm{a}_t)$ \cite{chua2018deep} at each time step $t$; \textbf{the prior probability $p_{D}(\theta)$} is derived from the posterior probability $p(\theta|D)$, where the parameter $\theta$ is inferred from the training dataset $D$; and lastly, \textbf{the optimality likelihood} $p(\mathcal{O}|\tau)$ is defined in relation to the expected reward and utility formulation of several trajectories, expressed as $\mathcal{F}_r\cdot p\left(\mathcal{O}_r|\tau\right):=\widetilde{r}(\tau)$ and $\mathcal{F}_g\cdot p\left(\mathcal{O}_{g_i}|\tau\right):=\widetilde{g}_{i}(\tau)$ , where the operator family $\mathcal{F}=\left\{\mathcal{F}_r,\mathcal{F}_g\right\}$ and the optimality family $\mathcal{O}=\{\mathcal{O}_{r},\mathcal{O}_{g_i}\}$ establish this relationship. In \textbf{Section \ref{Section_WVI}} and \textbf{Section \ref{Section_formal}}, we offer theoretical understanding to illustrate how such specific definitions influence the RL's global convergence.

\textbf{Constrained Reinforcement Learning as Probabilistic Graphical Models.} \; Specifically, we consider a Constrained Markov Decision Process (CMDP) \cite{altman1999constrained}, a formal framework for constrained RL, which is formulated as a discounted Markov decision process with additional constrained objectives, i.e., a tuple $\left \langle S,A,P,R,G,\gamma \right \rangle$: $S$ is a finite set of states ${ \left\{ \bm{s} \right\}}$; $A$ is a finite set of actions ${ \left\{ \bm{a} \right\}}$; $P: S\times A\rightarrow S$ is a finite set of transition probabilities ${ \left\{ p(\bm{s'}|\bm{s},\bm{a}) \right\}}$; $R: S\times A\times S \rightarrow \mathbb{R}$ is a finite set of bounded immediate rewards ${ \left\{ r \right\}}$; $G: S\times A\times S \rightarrow \mathbb{R}$ comprises a finite collection of unity functions ${ \left\{g\right\}}$, where, upon satisfying the expected constraints $g_i$, the unity-optimality variable is specified as $\mathcal{O}_{g_i}=1$; and $\gamma\in[0,1]$ is the discount rate. A CMDP is presented as:
\begin{equation}
\underset{\pi}{\rm{max}}\: \mathcal{J}_r(\pi), \quad {\rm{s.t.}}\quad \mathcal{J}_{g,i}(\pi)\leq \bm{b}_i+\bm{\tau}_{c},\quad i=1,...,n
\label{constrained_rl_problem}
\end{equation}
where $\mathcal{J}_r(\pi):= {\mathbb{E}}[{\sum\limits_{t=0}^{\infty}\gamma^{t}}r(\bm{s}_t,\bm{a}_t)|\pi, {\bm{s}_0}=s]$ and $\mathcal{J}_{g,i}(\pi):= {\mathbb{E}}[{\sum\limits_{t=0}^{\infty}\gamma^{t}}g_{i}(\bm{s}_t,\bm{a}_t)|\pi, {\bm{s}_0}=s]$ are the value function associated with the immediate reward $r$ and the utility $g$, respectively; $b_i$ is a fixed limit for the $i$-th constraint; and $\bm{\tau}_{c}$ is the tolerance. \autoref{pgm_saferl} shows how constrained RL can be viewed as a novel variation of PGMs. A complete table of nomenclature is provided in \textbf{Appendix \ref{section_notation_table}} for convenient reference.

\section{Method: Adaptive Sliced Wasserstein Variational Optimization (AWaVO)}
\label{alg_sum}
\begin{figure*}[ht]
  \centering
  \includegraphics[scale=0.57]{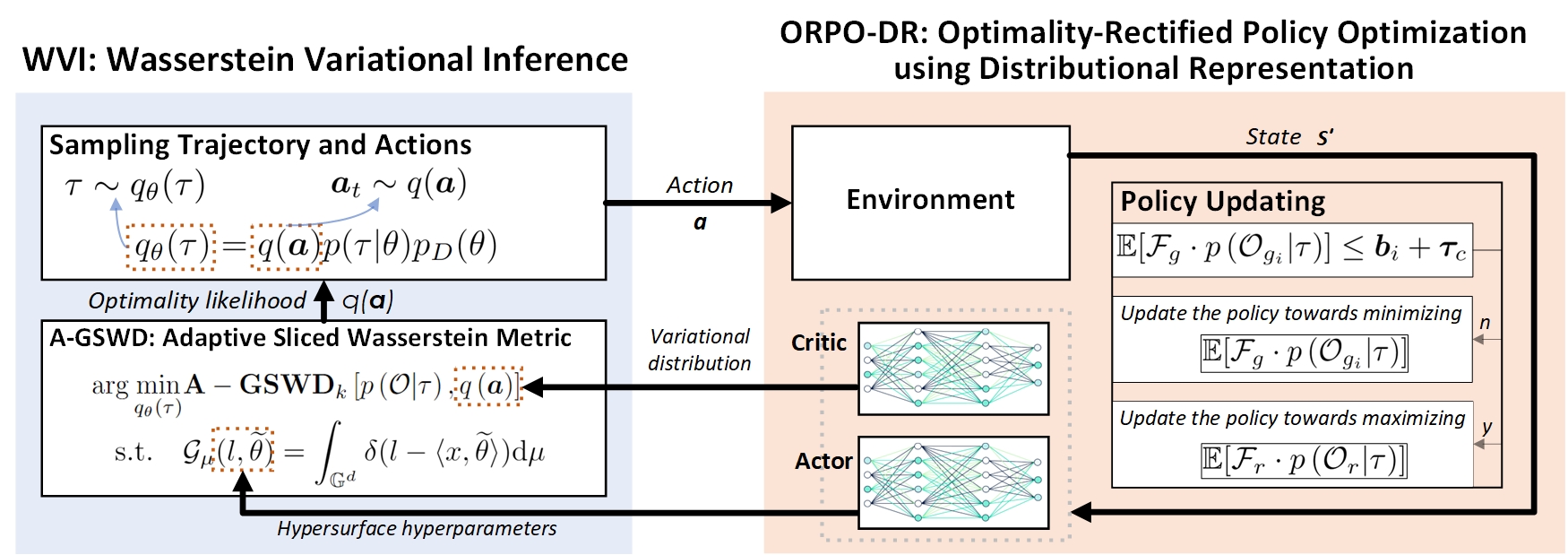}
  \caption{The algorithmic framework of AWaVO. We reform constrained RL as a Wasserstein variational optimization setup, consisting of two primary submodules: ORPO-DR and WVI (\textbf{Section \ref{alg_sum}}).}
  \label{Adaptive_slicing_framework}
\end{figure*}

In this section, we present AWaVO's two primary submodules: WVI and ORPO-DR. The detailed algorithm is outlined in Algorithm~{\autoref{AWaVO}}, and the overarching algorithmic structure is depicted in \autoref{Adaptive_slicing_framework}.

\subsection{WVI: Wasserstein Variational Inference}
\label{Section_WVI}
\textbf{Variational Inference for Dynamic Uncertainties.} \; Given uncertainties in a dynamics model, it is reasonable to assume that the optimal trajectories $\left\{\tau\right\}$ are uncertain. To infer optimal policies under uncertainties, let us consider a variational inference: $D(q_\theta(\tau)||p(\tau|\mathcal{O}))$, where, for simplicity, we use $p(\tau|\mathcal{O})$ to represent $p(\tau|\mathcal{O}_{t:T}=1)$ ; and $D(\cdot||\cdot)$ represents a distance metric between two probabilities. Building upon \autoref{Bayes_pro}, the variational distribution $q_\theta(\tau)$ is constructed as $q_\theta(\tau)=q(\bm{a})p(\tau|\theta)p_D(\theta)$. The construction suggests an assumption that the state transitions are controlled by $p(\bm{s}_{t+1}|\bm{s}_t,\bm{a}_t)$. According to \autoref{Bayes_pro}, we formulate the posterior as $p(\tau|\mathcal{O})\propto p(\mathcal{O}|\tau)p(\tau|\theta)p_D(\theta)$ (see \textbf{\cref{assumption_2}} for our implementation details).

While Kullback-Leibler (KL) divergence is widely used in conventional variational inference, its application in certain practical implementations can be risky due to its limitations, including asymmetry and infinity, arising when there are unequal supports. In this section, we extend the Wasserstein distance into the variational inference, and present the derivation of how we transform the GSWD between the two posteriors to the optimality likelihood $p(\mathcal{O}|\tau)$ and its approximation $q(\bm{a})$.

\textbf{Adaptive Generalized Sliced Wasserstein Distance.} \; GSWD has exhibited high projection efficiency in previous studies \cite{kolouri2019generalized,chen2021augmented} (please refer to \textbf{Appendix \ref{Section_bg_WD}} for a comprehensive background and definition of Wasserstein distance). However, the identification of the hypersurface hyperparameters, such as $l$ and $\widetilde{\theta}$, remains to be a challenge. The selection of these parameters, specifying the hypersurface along with its slicing direction, is generally a task-specific problem and requires prior knowledge or domain expertise. We now present a new adaptive sliced Wasserstein distance, called A-GSWD, that integrates GSWD with ORPO-DR, an adaptive process for determining the parameters of a hypersurface.

\begin{definition}
\label{a_gswd_def}
Given SWD and GSWD (defined in \textbf{Appendix \ref{Section_swd}} and \textbf{Appendix \ref{Section_gswd}}, respectively), we define A-GSWD by utilizing ORPO-DR for the adaptive slicing:
\begin{equation}
\begin{aligned}
    \bf{A}-\bf{GSWD}_{k}(&\mu,\nu) =\\
    &\left(\int_{\mathcal{R}_{\widetilde{\theta}}} W_k^{k}\left(\mathcal{A}_{\mu}\left(\cdot,\widetilde{\theta}\right), \mathcal{A} _{\nu}\left(\cdot,\widetilde{\theta}\right)\right) \mathrm{d}\widetilde{\theta}\right)^{\frac{1}{k}} \nonumber
\end{aligned}
\end{equation}
where $\mu,\nu\in P_k(\mathcal{X})$ are two measures of probability distributions over the space $\mathcal{X}$ (see \textbf{Appendix \ref{Section_swd}} for details). $l\in\mathbb{R}$ and $\widetilde{\theta}\in\mathcal{R}_{\widetilde{\theta}}$ represent the parameters of hypersurfaces, both of which are the outputs from actor networks in ORPO-DR (see \autoref{Adaptive_slicing_framework} for details). The Adaptive Generalized Radon Transforms (AGRT) $\mathcal{A}$ is used as a push-forward operator $\mathcal{A}_{\mu}$, defined by $\mathcal{A}_{\mu}(l,\widetilde{\theta}) = \int_{\mathbb{G}^d} \delta(l-\alpha( x,\widetilde{\theta} )) \mathrm{d}\mu$, where $\alpha(x,\widetilde{\theta})$ is \textit{a defining function} satisfying the conditions \textbf{H.1-H.4} in \cite{kolouri2019generalized}. $\mathcal{R}_{\widetilde{\theta}}\subset\mathbb{R}^d$ is a compact set of all feasible parameters $\widetilde{\theta}$, where $\mathcal{R}_{\widetilde{\theta}}=\mathbb{S}^{d-1}$ for $\alpha(\cdot,\widetilde{\theta})=\langle \cdot,\widetilde{\theta} \rangle$.
\end{definition}

Although the proposed adaptive slicing method, i.e., A-GSWD, improves the efficiency and accuracy of the Wasserstein distance computation, its demonstration on a valid metric guarantee remains a problem \cite{kolouri2019generalized}. In \textbf{Section \ref{Section_formal}}, we prove that the proposed A-GSWD is a true metric that satisfies non-negativity, symmetry, the triangle inequality and ${\bf{A-GSWD}}_{k}(\mu,\mu)=0$, respectively.

We then employ A-GSWD to address the variational inference, i.e., minimizing the distance $D(q_\theta(\tau)||p(\tau|\mathcal{O}))={\bf{A-GSWD}}_{k}(q_\theta(\tau),p(\tau|\mathcal{O}))$ between the variational distribution $q_\theta(\tau)$ and the posterior distribution $p(\tau|\mathcal{O})$. Subsequently, the variational inference can be reformulated to the minimization problem, as shown in WVI of \autoref{Adaptive_slicing_framework}: $\arg \underset{q_\theta(\tau)}{\rm{min}}{\bf{A-GSWD}}_{k}\left(q\left(\bm{a}\right),p\left(\mathcal{O}|\tau\right)\right)$, where $p(\mathcal{O}|\tau)$ represents the optimality likelihood, and the detailed derivation is in \textbf{Appendix \ref{Subsection_object}}.

\subsection{ORPO-DR: Optimality-Rectified Policy Optimization using Distributional Representation}
\label{Subsec_orpodr}
The current policy optimization for constrained RL can be classified into two categories: primal-dual and primal approaches \cite{xu2021crpo}. The former, transforming the constrained problem into an unconstrained one, are most commonly used although sensitive to Lagrange multipliers and other hyperparameters, such as the learning rate. On the other hand, the latter (i.e., primal approaches) require less hyperparameter tuning but have received less attention in terms of convergence demonstration compared to the primal-dual approaches.

\textbf{Policy Optimization combining Optimality Likelihood.} \quad Based on \textbf{Section \ref{Section_WVI}}, a constrained RL problem, as outlined in \autoref{constrained_rl_problem}, can be iteratively substituted and resolved as:
\begin{equation}
\nonumber
\left\{
\begin{aligned} 
&\arg \underset{q(\bm{a})}{\rm{max}}\mathbb{E}[\mathcal{F}_r \cdot p\left(\mathcal{O}_r|\tau\right)], \;{\mathbb{E}[\mathcal{F}_g \cdot p\left(\mathcal{O}_{g_i}|\tau\right)]}\leq \bm{b}_i+\bm{\tau}_{c} \\
&\arg \underset{q(\bm{a})}{\rm{min}}\mathbb{E}[\mathcal{F}_g\cdot p\left(\mathcal{O}_{g_i}|\tau\right)], \;\;otherwise
\end{aligned}
\right.
\end{equation}
where we recall that $\left\{\mathcal{F}_r,\mathcal{F}_g\right\}$ are two operators defined as $\mathcal{F}_r\cdot p\left(\mathcal{O}_r|\tau\right):=\widetilde{r}(\tau)$ and $\mathcal{F}_g\cdot p\left(\mathcal{O}_{g_i}|\tau\right):=\widetilde{g}_{i}(\tau)$, respectively. Furthermore, we can calculate the accumulated reward and utility function: $\widetilde{r}(\tau)={\mathbb{E}}[{\sum\limits_{t=0}^{T-1}\gamma^{t}}r(\bm{s}_t,\bm{a}_t)]$ and $\widetilde{g}_{i}(\tau)= {\mathbb{E}}[{\sum\limits_{t=0}^{T-1}\gamma^{t}}g_{i}(\bm{s}_t,\bm{a}_t)]$. Consequently, we obtain $\mathcal{J}_r(\pi)={\mathbb{E}}[\widetilde{r}(\tau)]$ and $\mathcal{J}_{g,i}(\pi)={\mathbb{E}}[\widetilde{g}_{i}(\tau)]$ if $T=\infty$.

If we only define $\mathcal{F}_r\propto \log[\cdot]$, it becomes equivalent to the formulation used in \cite{levine2018reinforcement,okada2020variational,okada2018acceleration}. In this case, we can retrieve an optimization process that resembles Model Predictive Path Integral (MPPI) \cite{okada2018acceleration}. The design of reward functions in the traditional RL is typically based on task-specific heuristics which is often considered as much an art as science. We will present such interpretation in \textbf{Section \ref{Section_formal}} to show how the reward operator family $\mathcal{F}$ acts on convergence, as well as a more rigorous approach to ensure guaranteed global convergence rate during the training process. Additionally, in \textbf{Section \ref{Section_8}}, we empirically verify these theoretical guarantees.

\textbf{Policy Updating.} \quad As shown in Algorithm~{\autoref{Opt_Rect_Policy_Update}}, we first update the policy towards either maximizing $\mathbb{E}[\mathcal{F}_r \cdot p\left(\mathcal{O}_r|\tau\right)]$ or minimizing $\mathbb{E}[\mathcal{F}_g\cdot p\left(\mathcal{O}_{g_i}|\tau\right)]$ by using the distributional representation (introduced in \textbf{\autoref{Section_dist_rep}}), where the gradient of actor and critic network, denoted as $\delta_{\theta^\mu}$ and $\delta_{\theta^Q}$, are defined in \autoref{grad_AC} in \textbf{\autoref{Section_dist_rep}}. Then, as shown in ORPO-DR of \autoref{Adaptive_slicing_framework}, the actor network outputs the parameters to dynamically determine the hypersurfaces and the corresponding slicing directions
; and the critic network provides an entire state-action distribution, which is directly utilized as the variational distribution of the optimality likelihood $q(\bm{a})$ in A-GSWD, as shown in \autoref{Adaptive_slicing_framework}.

\begin{algorithm}[t]
\footnotesize
\caption{ORPO-DR: Optimality-Rectified Policy Optimization using Distributional Representation}
\hspace*{0.02in} {\bf Input:}
$\bm{s}_k$, $\bm{s}_{k+1}$, $\bm{\tau}_c$, $\theta^\mu$, $\theta^Q$\\
\hspace*{0.02in} {\bf Output:} updated $\theta^\mu$, $\theta^Q$
\begin{algorithmic}[1]
\STATE \textbf{Constraint Estimation}:\\ ${\mathcal{J}_{g,i}}(\pi_{\theta}(\bm{s},\bm{a}))=\mathbb{E}[\mathcal{F}_g\cdot p\left(\mathcal{O}_{g_i}|\tau\right)]$, $\ \forall i\in[1,p]$\\
\STATE \textbf{Policy Improvement}:\\
\IF{${\mathcal{J}_{g,i}}(\bm{\pi})\leq \bm{b}_i+\bm{\tau}_{c},\forall i\in[1,p]$}
\STATE update the policy towards maximizing $\mathbb{E}[\mathcal{F}_r \cdot p\left(\mathcal{O}_r|\tau\right)]$:\\ $\theta^\mu \gets \theta^\mu + l_{\mu}\delta_{\theta^\mu}$, and $\theta^Q \gets \theta^Q + l_{\theta} \delta_{\theta^Q}$\\
\ELSE
\STATE update the policy towards minimizing $\mathbb{E}[\mathcal{F}_g\cdot p\left(\mathcal{O}_{g_i}|\tau\right)]$:\\ $\theta^\mu \gets \theta^\mu - l_{\mu}\nabla_{\theta^\mu}\widetilde{g}_{i}(\tau)$, and $\theta^Q \gets \theta^Q - l_{Q} \nabla_{\theta^Q} \widetilde{g}_{i}(\tau)$
\ENDIF
\end{algorithmic}
\label{Opt_Rect_Policy_Update}
\end{algorithm}

\begin{algorithm}[t]
\footnotesize
\caption{AWaVO: Adaptive Sliced Wasserstein Variational Optimization}
\hspace*{0.02in} {\bf Input:}
$\bm{s}_k$, $\bm{s}_{k+1}$, $\theta^\mu$, $\theta^Q$\\
\hspace*{0.02in} {\bf Output:}
$\bm{a}_k$
\begin{algorithmic}[1]
\STATE \textbf{Initialize}:\\
    $\bm{\theta}=[\theta^\mu,\theta^Q]$: the parameters of actor and critic network\\
\REPEAT
    \FOR{$t=0,1,2,...,T-1$}
    \STATE \textbf{Parameter Identification:} achieve $p_{D}(\theta)$ by doing inference of the posteriors $p(\theta|D)$ (Section \ref{Section_seq_decision})\\
    \STATE \textbf{Policy Updating:} $\left\{\theta^\mu,\theta^Q\right\} \gets$ Exec. Algorithm~{\autoref{Opt_Rect_Policy_Update}} $(\bm{s}_k$, $\bm{s}_{k+1}$, $\bm{\tau}_c$, $\theta^\mu$, $\theta^Q)$ \\
    \STATE \textbf{Inference Execution:} do inference of the posterior probability, as described in Section \ref{Section_WVI}:\\$p(\tau|\mathcal{O}_{t:T}) \gets\arg \underset{q_\theta(\tau)}{\rm{min}}{\bf{A-GSWD}}_{k}\left(q\left(\bm{a}\right),p\left(\mathcal{O}|\tau\right)\right)$\\
    \STATE sample actions $\bm{a}_k\gets p(\tau|\mathcal{O}_{t:T})$, execute $\bm{a}_k$, and observe $\bm{s}_{k+1}$\\
    \ENDFOR
\UNTIL{convergence}
\end{algorithmic}
\label{AWaVO}
\end{algorithm}

\begin{figure*}[ht]
\subfigure[Acrobot tasks in OpenAI Gym]{
	\begin{minipage}[]{0.49\linewidth}
		\centering
		\includegraphics[scale=0.175]{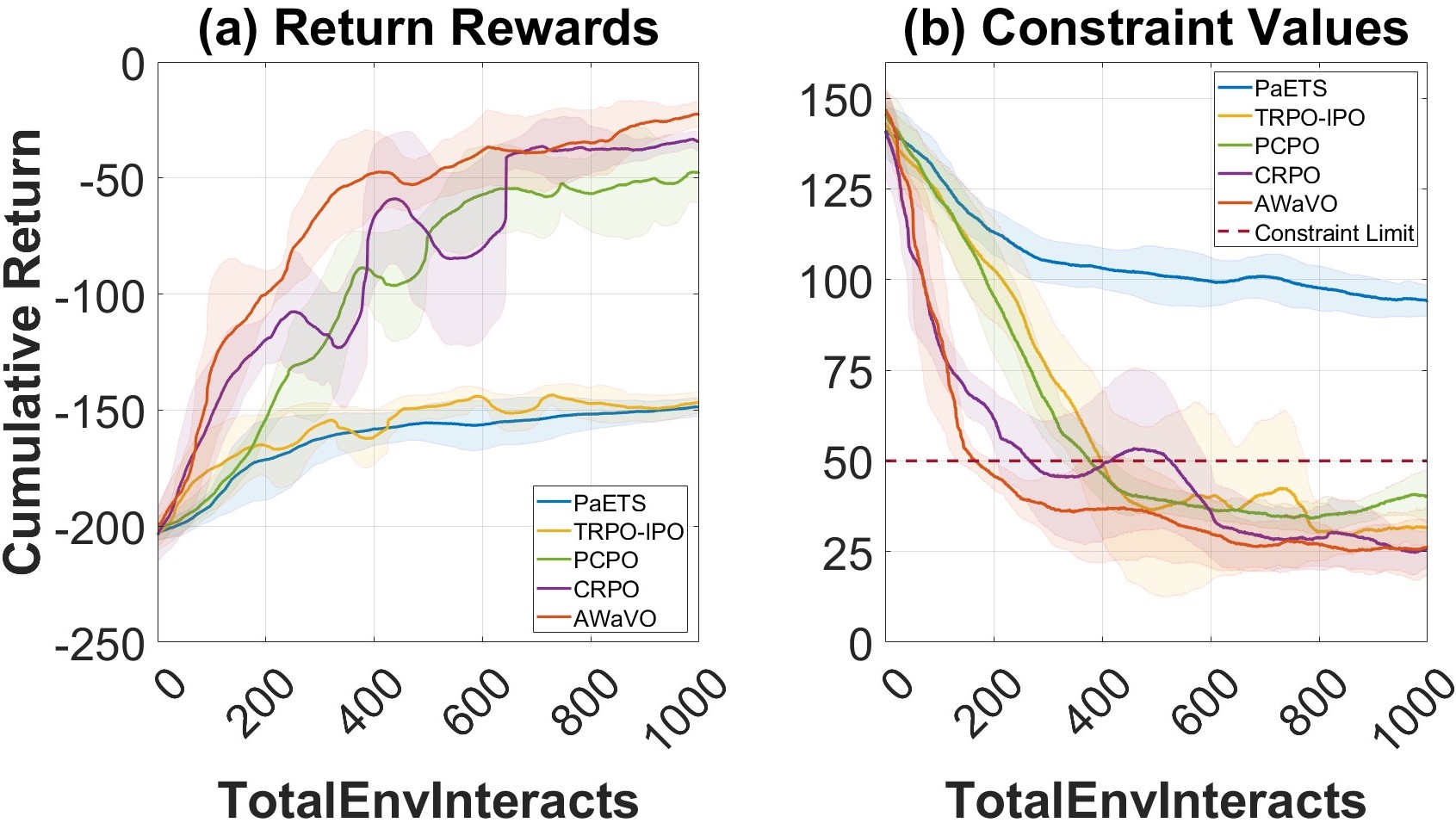}
		\label{learning_cur_Acrobot}
	\end{minipage}
 }
\subfigure[Cartpole tasks in OpenAI Gym]{
	\begin{minipage}[]{0.48\linewidth}
		\centering
		\includegraphics[scale=0.175]{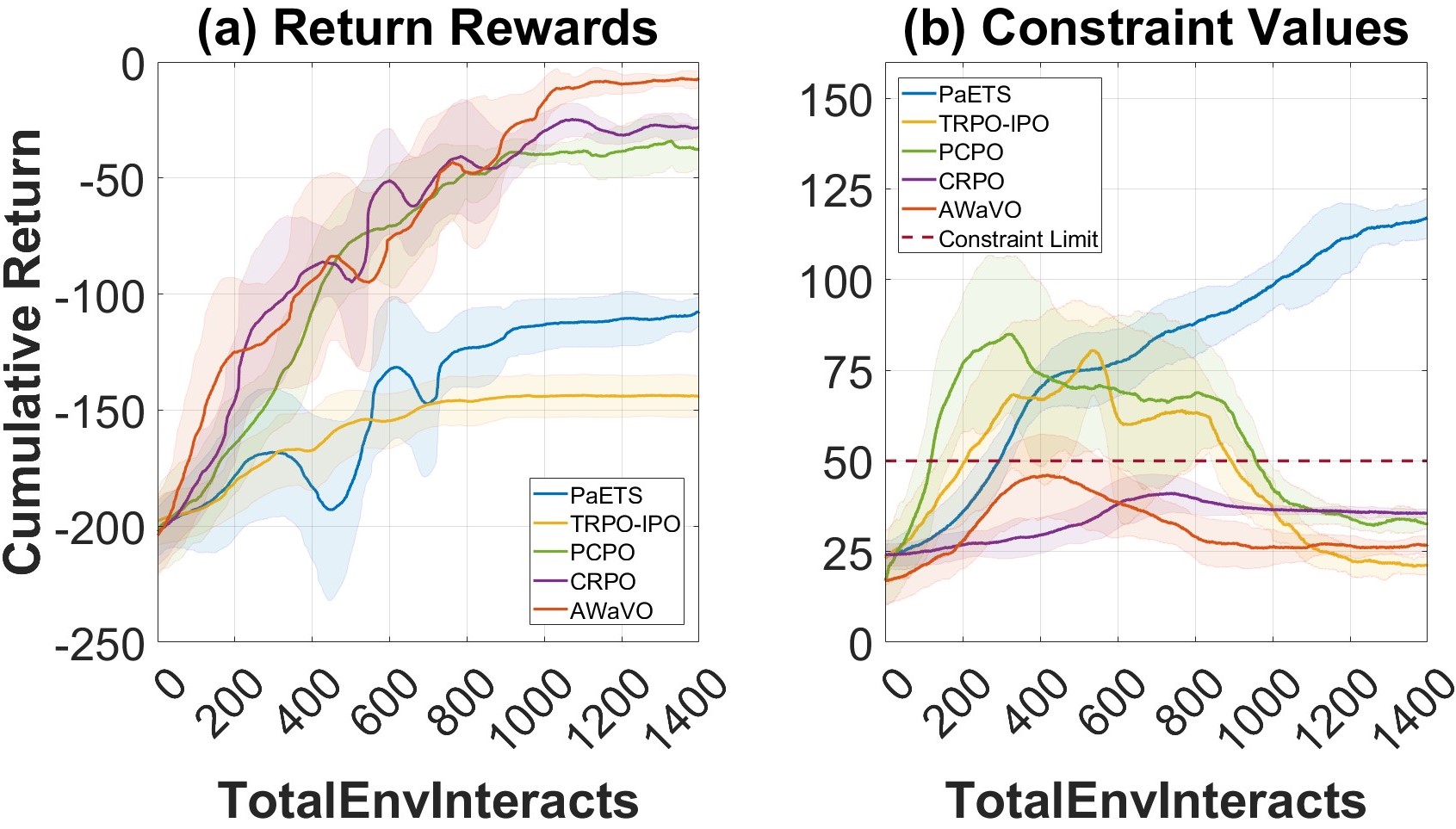}
		\label{learning_cur_cartpole}
	\end{minipage}
 }
\subfigure[Walker tasks in GUARD]{
	\begin{minipage}[]{0.49\linewidth}
		\centering
		\includegraphics[scale=0.175]{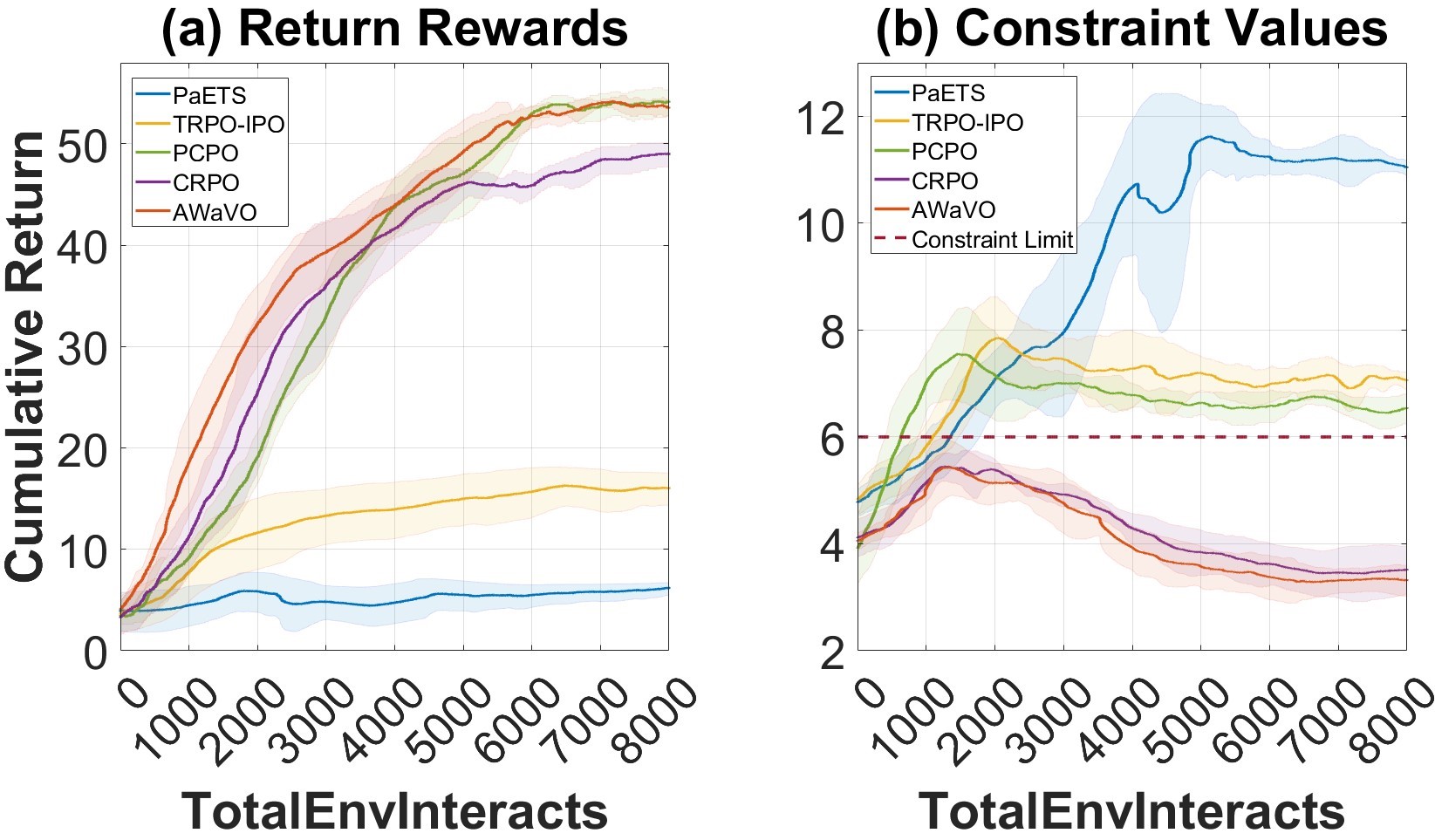}
		\label{learning_cur_Walker}
	\end{minipage}
}
\subfigure[Drone tasks in GUARD]{
	\begin{minipage}[]{0.48\linewidth}
		\centering
		\includegraphics[scale=0.175]{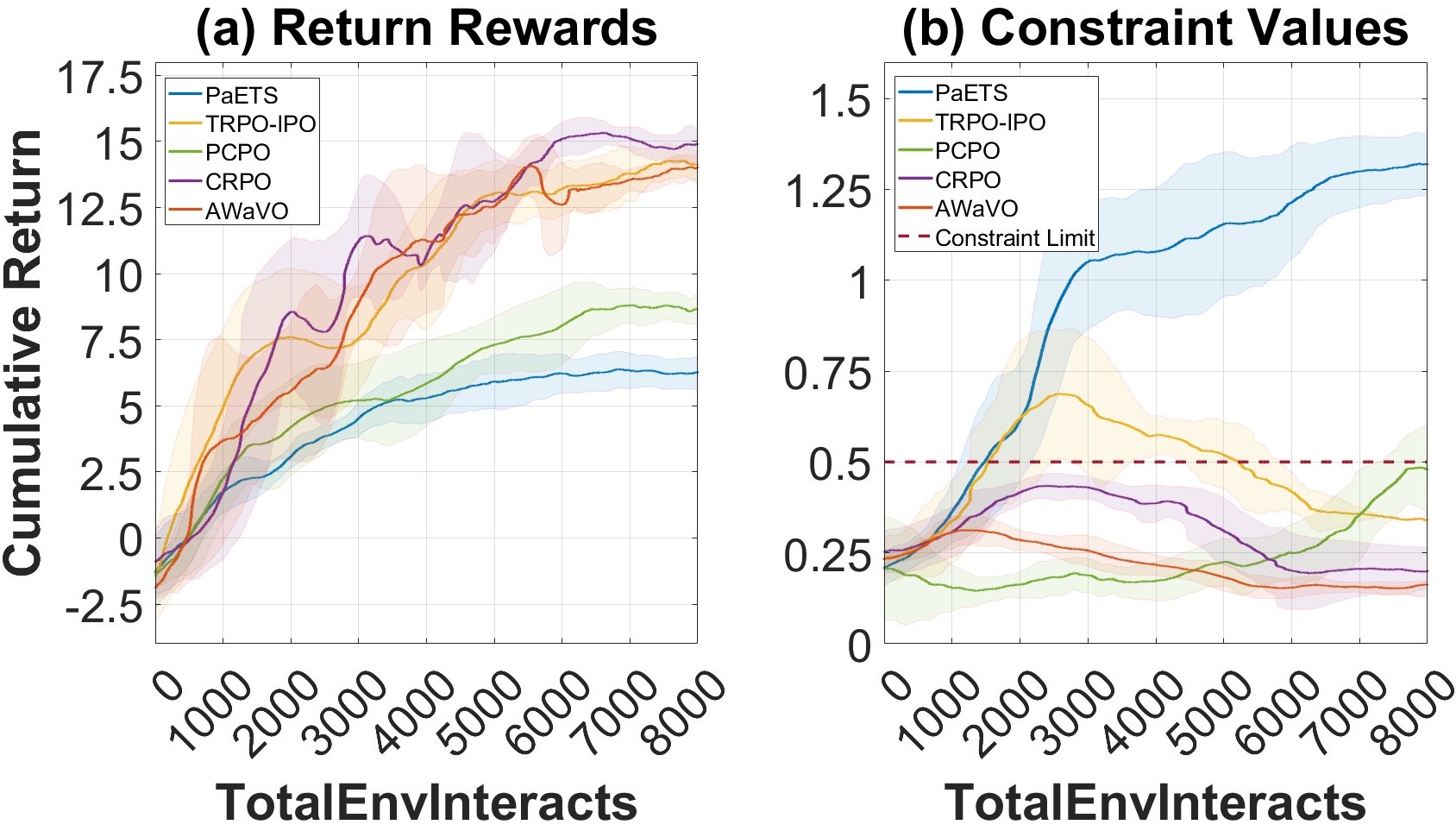}
		\label{learning_cur_drone}
	\end{minipage}
}
    \caption{Performance comparison over 10 seeds. CRPO and AWaVO outperform PaETS, with a trade-off highlighted: although PaETS offers probabilistic interpretation with Bayesian networks, its convergence is generally unstable. Our proposed AWaVO achieves a better balance between high performance and interpretability. In contrast to two other constrained RL algorithms, i.e., TRPO-IPO and PCPO, we observe an interesting result: PCPO performs better in tasks like Acrobot, Cartpole, and Walker, while TRPO-IPO outperforms PCPO in the more complex drone tasks (Figure~\autoref{learning_cur_drone}). Further, in \autoref{real_flight_tasks_conv}, we will explore more complex real-world tasks using an aerial robot.}
    \label{acro_cart_learning_curve}
\end{figure*}

\section{Formal Methods for Interpretability}
\label{Section_formal}
\begin{proposition}
\label{proposition_1}
\textit{(Pseudo-metric)}: Given two probability measures $\mu,\nu\in P_k(\mathcal{X})$ and a mapping $\alpha: \mathcal{X}\rightarrow\mathcal{R}_{\widetilde{\theta}}$, the adaptive slicing A-GSWD, defined in \textbf{\cref{a_gswd_def}}, with order $k$ in the range $[1,\infty)$, is a pseudo-metric that satisfies non-negativity, symmetry, the triangle inequality and ${\bf{A-GSWD}}_{k}(\mu,\mu)=0$. See \textbf{Appendix \ref{Section_proof}} for Proof.
\end{proposition}

\begin{remark}
\label{remark}
The adaptive slicing A-GSWD, with order $k\in[1,\infty)$, is a true metric \textit{if and only if} the AGRT $\mathcal{A}$, defined in \textbf{\cref{a_gswd_def}}, is an injective mapping.
\end{remark}

We make the following three assumptions.
\begin{assumption}
\label{assumption_1}
We define the function:
$p(\tau|\mathcal{O})=p(\mathcal{O}|\tau)p(\tau|\theta)p_D(\theta)$.
\end{assumption}

\begin{assumption}
\label{assumption_2}
Let $\Psi(s,a)$ be a feature vector, and $\chi_\pi$ be a stationary distribution in CMDP: $(s,a)\sim\chi_\pi$. There exists a constant $\hat{C_0}$ such that for any $\varrho\geq0$, it holds that $p(\left|x^\mathsf{T} \Psi(s,a)\right|\leq\varrho)\leq \hat{C_0}\cdot\varrho$, where $x\in\mathbb{R}^d$.
\end{assumption}

\begin{assumption}
\label{assumption_3}
We define the family of functions:
\begin{equation}
\nonumber
\begin{aligned}
\mathcal{B}_{R,\infty}&=f((s,a);\theta_q)=f((s,a);\theta_{q,0})\\
&+\int\mathbf{1}({\theta_q}^\mathsf{T}\Psi(s,a)>0)\cdot\omega(\theta_q)^\mathsf{T}\Psi(s,a)\mathrm{d}\varphi(\theta_q)
\end{aligned}
\end{equation}
where $f((s,a);\theta_q)$ is an $H$-layer neural network corresponding to the initial parameter $\theta_{q,0}$. The weighted function $\omega(\theta_q): \mathbb{R}^d\rightarrow\mathbb{R}^d$ satisfies $\left \| \omega(\cdot)\right \|_\infty\leq C_R/ \sqrt{d}$, where $C_R\in \mathbb{R}$ denotes an upper bounded value and $d\geq2$. $\varphi(\cdot):\mathbb{R}^d\rightarrow\mathbb{R}$ represents the density of the weight distribution. We assume $Q^\pi\in\mathcal{B}_{R,\infty}$, for all $\pi$. 
\end{assumption}

\textbf{\cref{assumption_1}} specifies that in our implementation, the term $p(\tau|\mathcal{O})$ is explicitly formulated as $p(\mathcal{O}|\tau)p(\tau|\theta)p_D(\theta)$. \textbf{\cref{assumption_2}} implies that the probability density of the distribution $\Psi(s,a)$ is uniformly upper-bounded over the unit sphere, a condition achievable in most ergodic Markov chains \cite{mitrophanov2005sensitivity,xu2021crpo}. 
\textbf{\cref{assumption_3}} implies a mild and broadly applicable regularity condition on $Q^\pi$, as $\mathcal{B}_{R,\infty}$ can be interpreted as a function class with infinite width neural networks, thus representing a sufficiently general set of functions.

To establish a link between the reward operator family $\mathcal{F}$ and the global convergence of ORPO-DR, here, we introduce \textbf{\cref{conditions}} and then present \textbf{\cref{theorem_1}} \textit{(Global Convergence)}.

\begin{conditions}
\label{conditions}
The reward operator family $\mathcal{F}=\left\{\mathcal{F}_r,\mathcal{F}_g\right\}$ satisfies that: (\textit{i}) $\mathcal{F}_r$ is monotonically increasing and continuously defined on $(0,1]$, and the range covers $[r_{\min}, r_{\max}]$;
and (\textit{ii}) $\mathcal{F}_g$ is monotonically decreasing and continuously defined on $(0,1]$, and the range covers $[r_{\min}, r_{\max}]$.
\end{conditions}

\begin{theorem}
\label{theorem_1}
\textit{(Global Convergence)}: Given the policy in the $i$-th policy improvement $\bm{\pi}^{\bm{i}}$, $\bm{\pi}^{\bm{i}} \rightarrow \bm{\pi}^{*}$ and $i\rightarrow\infty$, suppose \cref{assumption_1} holds, there exists $Q^{\bm  {\pi}^{*}}(s, a) \geq Q^{\bm{\pi}^{\bm{i}}}(s, a)$ \textit{if and only if} the reward operator family $\mathcal{F}$ satisfies the both \textbf{\cref{conditions}}. See \textbf{Appendix \ref{Section_proof}} for Proof.
\end{theorem}

Subsequently, we present a more rigorous comprehension of how $\mathcal{F}$ precisely influences the convergence rate. To our best knowledge, this study represents the first attempt to develop an intrinsic interpretation of how the reward function design influences convergence within the RL community.

\begin{theorem}
\label{theorem_2}
\textit{(Global Convergence Rate)}: Let $m$ and $H$ be the width and layers of a neural network, $K_{td}=(1-\gamma)^{-\frac{3}{2}}m^{\frac{H}{2}}$ be the iterations required for convergence of the distributional Temporal Difference (TD) learning (defined in \autoref{TD_learning}), $l_{Q}=\frac{1}{\sqrt{T}}$ be the policy update (in \textit{Line 4} of Algorithm~{\autoref{Opt_Rect_Policy_Update}}) and  $\bm{\tau}_{c}=\Theta(\frac{1}{(1-\gamma)\sqrt{T}})+\Theta(\frac{1}{(1-\gamma)Tm^{\frac{H}{4}}})$ be the tolerance (in \textit{Line 3} of Algorithm~{\autoref{Opt_Rect_Policy_Update}}). Suppose Assumptions 1-3 hold. There exists a global convergence rate of $\Theta(1/\sqrt{T})$, and a sublinear rate of $\Theta(1/\sqrt{T})$ if the constraints are violated with an error of $\Theta(1/{m^{\frac{H}{4}}})$, with probability of at least $1-\delta$. This holds \textit{if and only if} the reward operator family $\mathcal{F}$ satisfies \textbf{\cref{conditions}}. See \textbf{Appendix \ref{Section_proof}} for Proof.
\end{theorem}

\textbf{Probabilistic interpretation on sequential decisions.} \; We now quantitatively establish the relationships between latent factors, such as disturbances, that possibly influence decision-making and the sequential decisions, namely trajectories, by providing a probabilistic interpretation. Referring to the abbreviation presented in \autoref{Bayes_pro}, we reform it as: $p(\tau|D)=p(\mathcal{O}|\tau) \cdot p(\bm{s},\bm{a}|\theta) \cdot p_{D}(\theta)$. Then, the latent factors are denoted by $L=\left\{L_i\right\}_{i=0}^{M-1}$, where $M$ represents the total number of defined factors. By applying the chain rule to the posterior probability $p(\tau|D)$, we have $\left\{p(\tau|L_i)\right\}_{i=0}^{M}=\left\{\frac{p(\tau|D)}{p(L_i|D)}\right\}_{i=0}^{M}$, where
the equation provides a decomposition of the joint posterior probability $p(\tau|D)$ into conditional probabilities that involve individual factors $L_i$. This decomposition is not only notable for its theoretical simplicity but facilitates a practical probabilistic understanding of how each factor influences policy in real-world safety-critical scenarios, such as robot autonomy. In \textbf{Section \ref{Section_8}}, we showcase numerical examples to illustrate such probabilistic interpretation.

\begin{figure}[t]
    \centering
    \includegraphics[scale=0.51]{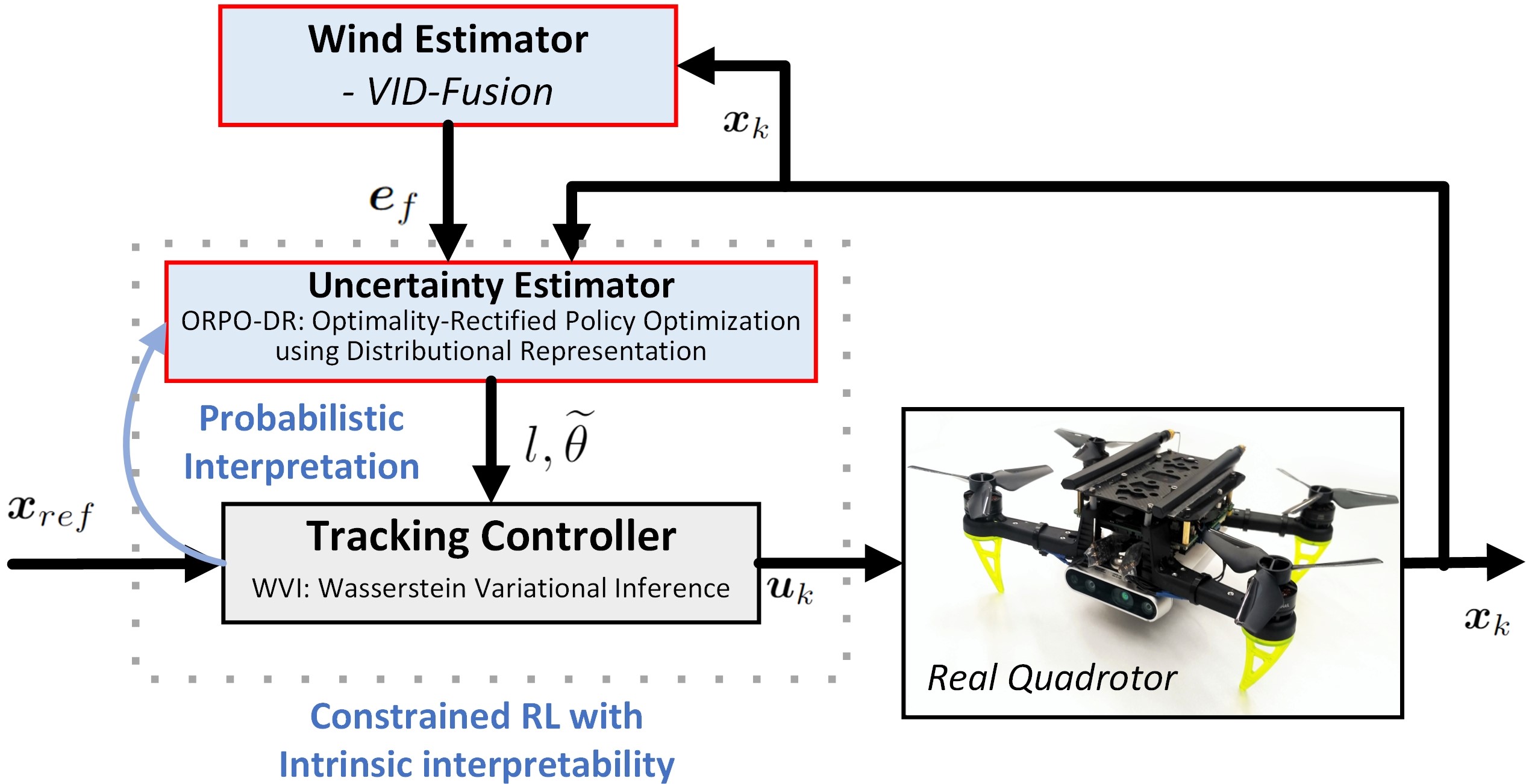}
    \caption{We use our AWaVO as the tracking controller for a quadrotor, where ORPO-DR is employed as the uncertainty estimator, and WVI using A-GSWD is leveraged as the controller.}
    \label{RL_Control_framework}
\end{figure}

\begin{figure}[t]
    \centering
    \includegraphics[scale=0.1795]{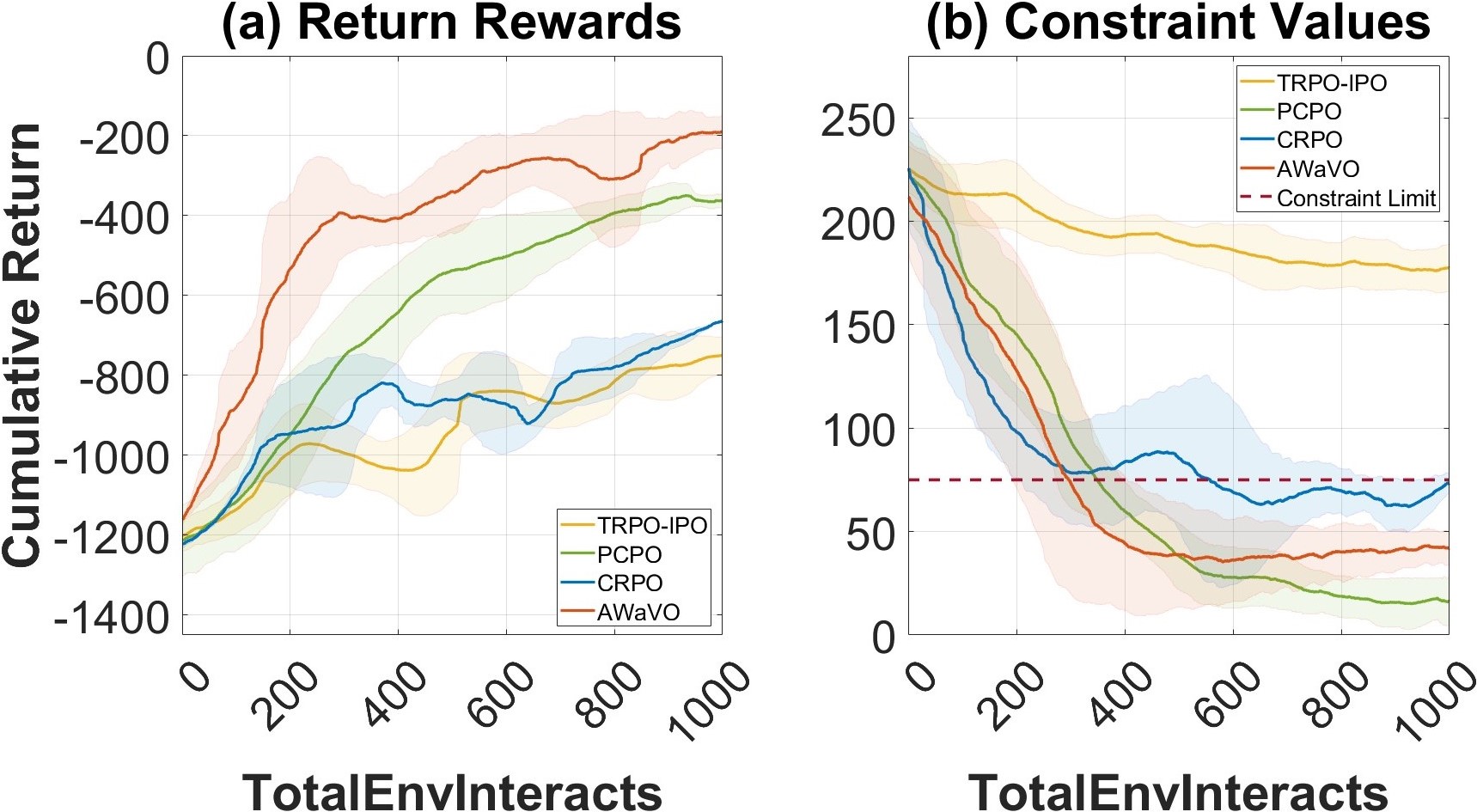}
    \caption{Performance comparison in a real quadrotor: our AWaVO slightly outperforms the constrained RL approach, i.e., PCPO, whilst achieving interpretability in \autoref{Pro_interpret}. }
    \label{real_flight_tasks_conv}
\end{figure}

\section{Experiments}
\label{Section_8}

\begin{figure*}[t]
\centering
\includegraphics[scale=0.14]{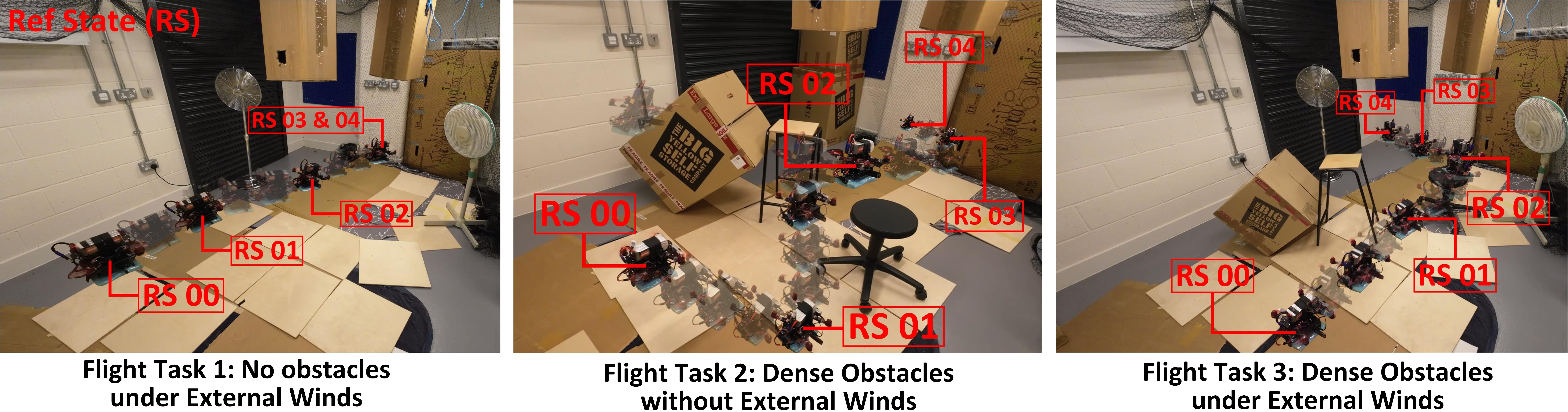}
\caption{Real quadrotor Flight Tasks (FTs): \textit{(i) FT 1 -} tracking reference trajectories under external forces without obstacles; \textit{(ii) FT 2 -} tracking trajectories around dense obstacles; and \textit{(iii) FT 3 -} tracking trajectories under external forces around dense obstacles.}
\label{real_flight_tasks}
\end{figure*}
In this section, we explain our empirical assessments of AWaVO's performance in simulated platforms and real-world robot tasks. Initially, we perform tasks with multiple constraints in OpenAI Gym framework \cite{brockman2016openai}. Then we showcase AWaVO's practicality through real quadrotor Flight Tasks (FTs) to provide a more comprehensive assessment of its performance. These evaluations serve a dual purpose: to validate AWaVO's performance; and, critically, to empirically demonstrate its quantitative interpretability. This interpretability includes confirming properties such as the guaranteed convergence rate as demonstrated in \textbf{\cref{theorem_2}} and the probabilistic decision interpretation discussed in \textbf{Section~\ref{Section_formal}} within the context of sequential decision-making tasks.

\textbf{Comparative Performance in Simulated Tasks.} \; We conduct tasks with multiple constraints in OpenAI Gym \cite{brockman2016openai} and GUARD \cite{zhao2023guard} (a constrained RL benchmark): Acrobot, Cartpole, Walker and Drone. We use four appropriate constrained RL benchmarks: PaETS \cite{okada2020variational}, i.e., a Bayesian RL combining with variational inference,  TRPO-IPO \cite{liu2020ipo}, i.e., an enhanced variant of TRPO-Lagrangian \cite{bohez2019value}, PCPO \cite{yang2020projection}, i.e., an advanced variant of CPO \cite{achiam2017constrained} and CRPO \cite{xu2021crpo}, i.e., a primal constrained RL approach.

The AWaVO parameter settings given in \autoref{model_parameters} of \textbf{Appendix \ref{Subsec_exp_set}} are based on selected benchmarks, i.e., CRPO \cite{xu2021crpo} and GUARD \cite{zhao2023guard}. According to our proposed~\textbf{\cref{proposition_1}} and the \textbf{Proposition 1} presented in \cite{kolouri2019generalized}, the defining function $\alpha(\cdot,\widetilde{\theta})$ can be defined as homogeneous polynomials, i.e., $\alpha(\cdot,\widetilde{\theta})=\sum_{\left| \kappa\right|=m}\widetilde{\theta}_{\kappa}x^{\kappa}$, where the defining function $\alpha$ is injective if the degree of the polynomial $m$ is odd. Thus we set $m=3$ based on \cite{kolouri2019generalized}. The comprehensive task descriptions are available in \textbf{Appendix \ref{Subsection_task}}.

AWaVO's training depicted in \autoref{acro_cart_learning_curve} initially corresponds to the benchmarks provided by CRPO \cite{xu2021crpo} and GUARD \cite{zhao2023guard}. The distinction in iterations lies in showcasing the entire convergence process across four discrete tasks. We establish the constraint limit to facilitate a straightforward comparison of constraint convergence; see~\textbf{Appendix \ref{constraint_limit_setting}} for additional details. The tolerance is set as $\bm{\tau}_{c}=0.5$, following CRPO \cite{xu2021crpo}. By analyzing the comparative training performances in \autoref{acro_cart_learning_curve}, we observe that the superiority of CRPO and AWaVO stems from their primal constrained RL nature, which involves training under constraints and ensuring global convergence rate. Although CRPO exhibits comparable or slightly better convergence performance than AWaVO, as evident in Figure~\autoref{learning_cur_drone}, we place greater emphasis on two other aspects: training convergence under uncertainties and decision-making interpretation. In \autoref{real_flight_tasks_conv} below, we provide comparative demonstrations in real robot tasks to showcase how AWaVO effectively balances a trade-off between performance and interpretability in a more complex sequential decision-making scenario.

Furthermore, we empirically verify the formal method \textbf{\cref{theorem_2}} \textit{(Global Convergence Rate)} on the convergence rate, and conclude that, based on the average performance, the convergence rate of AWaVO is in the range of $\Theta(1/\sqrt{T})< C_{rate}\leq \Theta(1/T^{1.2})$. According to the results shown in Figure~\autoref{learning_cur_drone}, CRPO performs better than our AWaVO in the simulated drone task, with the absence of disturbances. Subsequently, in \autoref{real_flight_tasks_conv} below, we further evaluate these approaches in a real-world physical environment characterized by varying uncertainties, leading to different outcomes. It is worth noting that our approach incorporates two optimizations for handling uncertainties: variational inference and policy updating. This combination reduces the frequency of policy updates whilst enhancing our ability to handle uncertainties. In the upcoming real robot task, we will introduce variable disturbances to demonstrate our capability to optimize policies under uncertain conditions.

\begin{figure}[t]
\centering
\includegraphics[scale=0.084]{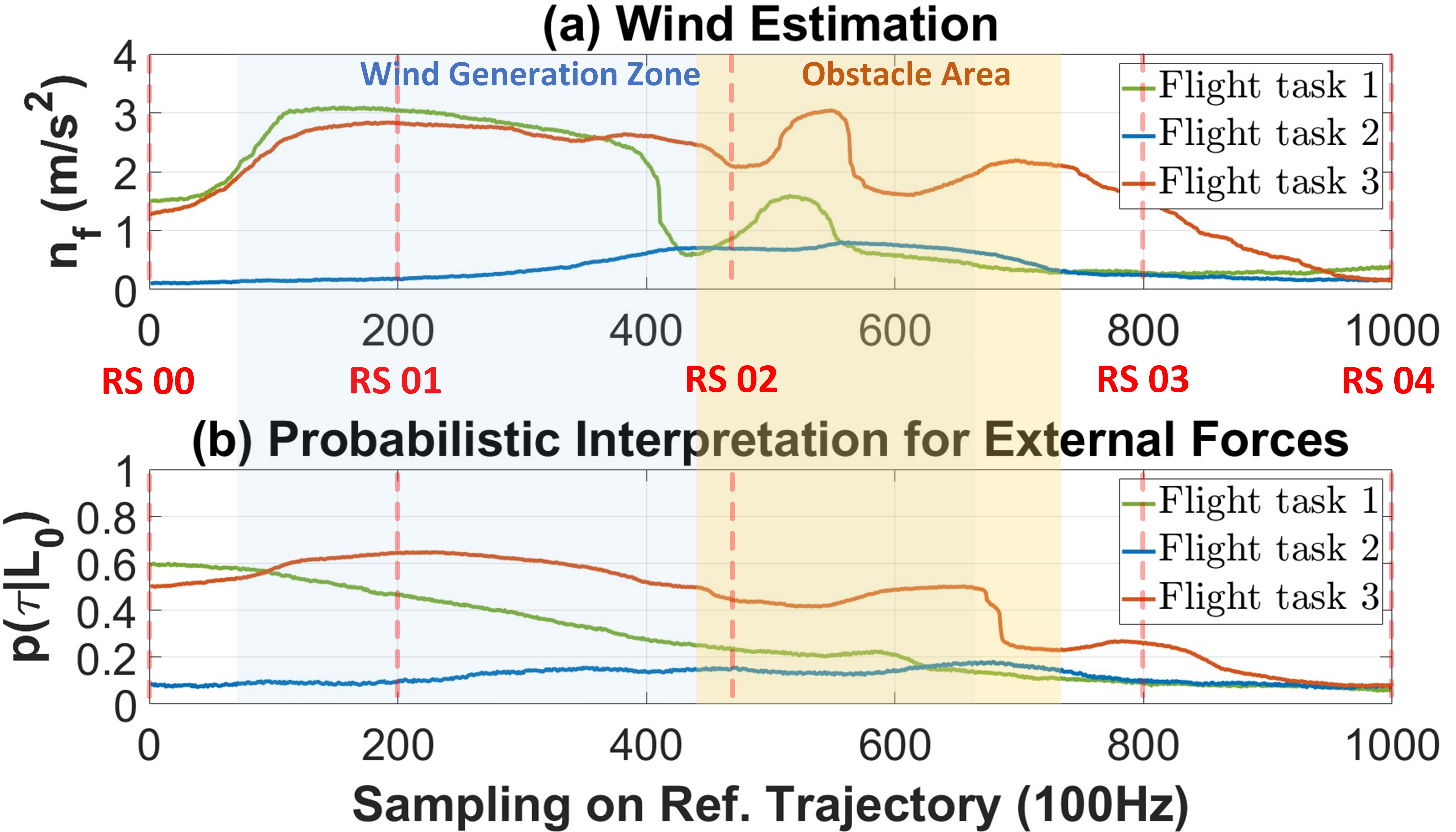}
\caption{Probabilistic interpretation of decisions: the probability $p(\tau|L_0)$ reveals the degree to which the measurement of external forces \cite{ding2021vid}, denoted as $n_f$, influences the decisions made by the quadrotor. For additional discussion on the case of `RS 02', as an example instance, please refer to \textbf{Appendix \ref{discuss_probabilistic_interpretation}}.}
\label{Pro_interpret}
\end{figure}

\textbf{Comparative Performance in Real-world Tasks.} \quad We demonstrate the effectiveness of AWaVO by practical implementation in real-world decision-making problems. The quadrotor's tracking control system, shown in \autoref{RL_Control_framework}, is an end-to-end learning-based framework. The technical specification of our quadrotor is shown in \autoref{hardware_spec} of \textbf{Appendix \ref{Subsec_exp_set}}. The training convergence is demonstrated in \autoref{real_flight_tasks_conv}. The aim of the real-world tasks (shown in  \autoref{real_flight_tasks}) is to track the reference effectively and accurately, where VID-Fusion \cite{ding2021vid} is used to measure external forces such as aerodynamic effects. To view the hardware experiments in action, please refer to the accompanying video demonstration at \url{https://github.com/Alex-yanranwang/AWaVO}.

Next, we illustrate the interpretation of sequential decisions, i.e., the actual control commands fed into the four motors. Leveraging the Intel RealSense D435i depth camera onboard, we can detect obstacles and estimate external forces. These latent factors, denoted as $L={L_0, L_1}$, represent external forces and obstacles, respectively. The probability $p(\tau|L)$ reveals why the quadrotor makes these decisions and quantifies the extent to which factor $L$ contributes to the sequential decisions, i.e., the real-time trajectory $\tau$. \autoref{Pro_interpret} presents a quantitative interpretation, i.e., $p(\tau|L_0)$, indicating the magnitude and evolution that the external force $n_f$ impacts on the current control decisions. Please see \textbf{Appendix \ref{discuss_probabilistic_interpretation}} for further discussion on \autoref{Pro_interpret}.

Practically, this probabilistic interpretation represents significant progress in addressing a longstanding and challenging question: why do machine systems powered by Artificial Intelligence (AI) technologies make certain decisions, and what are the exact latent factors influencing those decisions? Such progress holds particular value for safety-critical industries like self-driving vehicles, aerospace engineering and high-frequency trading in financial services, particularly in cases where AI-based approaches exhibit erratic performance and thorough analysis is necessary.


\section{Limitation}
The primary limitation we encounter is ensuring the
trustworthiness of the posterior probability generated by the
critic network, which operates as a Bayesian network. Our
ongoing efforts involve applying statistical methods to establish a specific confidence interval for the Bayesian network’s
outcomes. Additionally, we intend to study AWaVO's suitability for more real-world safety-critical applications. For scalability, please refer to \textbf{Appendix \ref{scalability}}. This may serve as a starting point from which to formally analyze AWaVO's scalability and how the approach could extend to larger and more complex real-world domains. Another limitation is the lack of human-centered validation (although this would need to be taken on an \textit{ad hoc} basis per deployment context), and considering policy or regulatory issues.

\section{Conclusion}
Enthusiasm towards the possible applications of constrained RL is growing worldwide. The insufficient ability to interpret agent actions and policy optimizations, however, poses a significant hurdle in deploying RL in safety-critical domains like advanced manufacturing and financial trading. Our primary motivation in introducing AWaVO, an intrinsically interpretable RL framework, is to tackle key challenges concerning convergence guarantees, optimization transparency, and sequential-decision interpretation. Empirical results demonstrate that the proposed AWaVO balances a reasonable trade-off between high performance and quantitative interpretability in both simulation and real quadrotor tasks.

\newpage

\section*{Acknowledgments}
This work was partially supported by NSF-UKRI [grant number NE/T011467/1]; and the Engineering and Physical Sciences Research Council [grant number EP/X040518/1].

\section*{Impact Statement}
This paper presents work whose goal is to advance knowledge about interpretability and reproducibility in the RL community. There are many potential societal consequences of our work, none of which we feel must be specifically highlighted here. Additionally, we provide the video and additional demonstration to the real-world quadrotor experiments.

\bibliography{example_paper}
\bibliographystyle{icml2024}

\clearpage
\appendix
\onecolumn
{\Large{\textbf{Appendix}}}
\section{Notation Table}
\label{section_notation_table}
\begin{table}[H]
    \centering
    \caption{Main Notation Conventions}
    \label{notation_table}
    \begin{adjustbox}{width=0.9\textwidth}
    \begin{tabular}{l l}
    \toprule
    \textbf{Parameters} & \textbf{Definition} \\
    \hline
    $\boldsymbol{s}_t$ & A state $\boldsymbol{s}$ sampled at a discrete timestamp $t$. \\
    $\boldsymbol{a}_t$ & An action $\boldsymbol{a}$ sampled at a discrete timestamp $t$. \\
    \hdashline
    $\tau$ & A trajectory.\\
    $D$ & Observed training dataset.\\
    \hdashline
    $S$& A finite set of states ${ \{ \boldsymbol{s} \}}$.\\
    $A$& A finite set of actions ${ \{ \boldsymbol{a} \}}$.\\
    $P$ & A finite set of transition probabilities ${ \{ p \}}$. \\
    $R$ & A finite set of bounded immediate rewards ${\{ r \}}$. \\
    $G$ & A finite collection of unity functions ${ \{g\}}$.\\
    \hdashline
    $\mathcal{O}_t=\{\mathcal{O}_{r,t},\mathcal{O}_{g_i,t}\}$ & \makecell[l]{Binary variables of the optimality. \\$\mathcal{O}_{r,t}=1$ and $\mathcal{O}_{g_i,t}=1$ signify that the trajectory $\tau$ is optimized and compliant with\\ the constraints.}\\
    \hdashline
    $p_{D}(\theta)$ & The posterior probability $p(\theta|D)$. \\
    $p(\mathcal{O}|\tau)$ & The optimality likelihood. \\
    \hdashline
    $\widetilde{r}(\tau)$ & The accumulated reward $\widetilde{r}(\tau)={\mathbb{E}}[{\sum\limits_{t=0}^{T-1}\gamma^{t}}r(\boldsymbol{s}_t,\boldsymbol{a}_t)]$. \\
    $\widetilde{g}_{i}(\tau)$  & The utility function $\widetilde{g}_{i}(\tau)= {\mathbb{E}}[{\sum\limits_{t=0}^{T-1}\gamma^{t}}g_{i}(\boldsymbol{s}_t,\boldsymbol{a}_t)]$. \\
    \hdashline
    $\mathcal{F}=\{\mathcal{F}_r,\mathcal{F}_g\}$ & \makecell[l]{$\mathcal{F}_r$ and $\mathcal{F}_g$ are the operators, with specifically defined as $\mathcal{F}_r\cdot$$p(\mathcal{O}_r|\tau)$$:=\widetilde{r}(\tau)$ and\\ $\mathcal{F}_g\cdot p(\mathcal{O}_{g_i}|\tau):=\widetilde{g}_{i}(\tau)$.} \\
    \hdashline
    $b_i$ & A fixed limit for the $i$-th constraint. \\
    $\boldsymbol{\tau}_{c}$ & The tolerance.\\
    \hdashline
    $l$ & A curve on the hypersurface in the spatial Radon transform. \\
    $\widetilde{\theta}$ & An unit vector tangent to $l$. Both $l$ and $\widetilde{\theta}$ represent the parameters of hypersurfaces. \\
    \hdashline
    $q_\theta(\tau)$ & A variational distribution. \\
    $q(\boldsymbol{a})$ & An approximation of $p(\mathcal{O}|\tau)$. \\
    \bottomrule
    \end{tabular}
    \end{adjustbox}
\end{table}

\section{Background on Wasserstein Distance}
\label{Section_bg_WD}
\subsection{Sliced Wasserstein Distance}
\label{Section_swd}
A fundamental challenge in both machine learning and statistics communities is to form effective metrics between pairs of probability distributions. Weaker notions, such as divergence measures, including KL divergence \cite{kullback1951information}, have been proposed and widely used. However, such measures do not satisfy the two basic properties of a metric, namely symmetry and triangle inequality. To address this issue, interest has rapidly increased in optimal transport in recent years. In this subsection, we introduce the Wasserstein distance and its variants, including SWD \cite{rabin2012wasserstein, nietert2022statistical} and GSWD \cite{kolouri2019generalized}, as metrics that conditionally satisfy the properties.

Let $\Gamma(\mu, \nu)$ be a set of all transportation plans $\gamma\in\Gamma(\mu, \nu)$, where $\gamma$ is a joint distribution over the space $\mathcal{X} \times \mathcal{X}$, and $\mu,\nu\in P_k(\mathcal{X})$ are two measures of probability distributions over $\mathcal{X}$. $P_k(\mathcal{X})$ represents a set of Borel probability measures with finite $k$-th moment on a Polish metric space \cite{villani2009optimal}. $d(x, y)$ represents a distance function over $\mathcal{X}$. The Wasserstein distance of order $k\in[1,\infty)$ between two measures $\mu,\nu$ is defined as \cite{villani2009optimal}: $W_k(\mu, \nu) = \left(\inf_{\gamma \in \Gamma(P, Q)} \int_{\mathcal{X} \times \mathcal{X}} d(x, y)^k \mathrm{d}\gamma(x, y) \right)^{1/k}$. This definition, however, involves solving an optimization problem that is computationally expensive in practical implementation, particularly for high-dimensional distributions. Thus sliced $k$-Wasserstein distance \cite{rabin2012wasserstein, nietert2022statistical}, defined over spaces of hyperplanes in $\mathbb{R}^d$, is proposed as a computationally efficient approximation: 
\begin{equation}
{\bf{SWD}}_{k}(\mu,\nu) = \left(\int_{\mathbb{S}^{d-1}} W_k^{k}\left(\mathcal{R}_{\mu}\left(\cdot,\widetilde{\theta}\right), \mathcal{R}_{\nu}\left(\cdot,\widetilde{\theta}\right)\right) \mathrm{d}\widetilde{\theta}\right)^{\frac{1}{k}}
\label{SWD_}
\end{equation}
where Radon transform $\mathcal{R}$ \cite{radon20051} is introduced in SWD to map a function $f(\cdot)$ to the hyperplanes $\left\{x\in \mathbb{R}^d|\langle x,\widetilde{\theta} \rangle=l\right\}$, i.e., $\mathcal{R}f(l,\widetilde{\theta}) = \int_{\mathbb{R}^d} f(x) \delta(l-\langle x,\widetilde{\theta} \rangle) \mathrm{d}x$: $l\in\mathbb{R}$ and $\widetilde{\theta}\in\mathbb{S}^{d-1}\subset\mathbb{R}^d$ represent the parameters of these hyperplanes. In the definition of SWD, the Radon transform $\mathcal{R}_{\mu}$ is employed as the push-forward operators, defined by $\mathcal{R}_{\mu}(l,\widetilde{\theta}) = \int_{\mathbb{R}^d} \delta(l-\langle x,\widetilde{\theta} \rangle) \mathrm{d}\mu$ \cite{kolouri2019generalized}.

\subsection{Generalized Sliced Wasserstein Distance}
\label{Section_gswd}
While SWD offers a computationally efficient way to approximate the Wasserstein distance, the projections are limited to linear subspaces, such as hyperplanes $\left\{x\right\}$. Due to the nature of these linear projections, the resulting metrics typically have low projection efficiency in high-dimensional spaces \cite{kolouri2019generalized,deshpande2019max}. Thus various variants of SWD are proposed to enhance its projection effectiveness. Specifically, the GSWD \cite{kolouri2019generalized}, defined in \autoref{generalized_SWD}, is proposed by incorporating nonlinear projections. Its main novelty is that Generalized Radon Transforms (GRTs) $\mathcal{G}$ \cite{beylkin1984inversion,ehrenpreis2003universality,homan2017injectivity}, i.e., $\mathcal{G}f(l,\widetilde{\theta}) = \int_{\mathbb{R}^d} f(x) \delta(l-\beta(x,\widetilde{\theta})) \mathrm{d}x$, are used to define the nonlinear projections towards hypersurfaces rather than linear projections to the hyperplanes in SWD. Let $\beta(x,\widetilde{\theta})$ be \textit{a defining function} when satisfying the conditions \textbf{H.1-H.4} in \cite{kolouri2019generalized}.
\begin{equation}
{\bf{GSWD}}_{k}(\mu,\nu) = \left(\int_{\mathcal{X}_{\widetilde{\theta}}} W_k^{k}\left(\mathcal{G}_{\mu}\left(\cdot,\widetilde{\theta}\right), \mathcal{G} _{\nu}\left(\cdot,\widetilde{\theta}\right)\right) \mathrm{d}\widetilde{\theta}\right)^{\frac{1}{k}}
\label{generalized_SWD}
\end{equation}

where $\widetilde{\theta}\in\mathcal{X}_{\widetilde{\theta}}$ and $\mathcal{X}_{\widetilde{\theta}}$ is a compact set of all feasible parameters $\widetilde{\theta}$ for $\beta(\cdot,\widetilde{\theta})$, e.g., $\mathcal{X}_{\widetilde{\theta}}=\mathbb{S}^{d-1}$ for $\beta(\cdot,\widetilde{\theta})=\langle \cdot,\widetilde{\theta} \rangle$. The GRT operator $\mathcal{G}_\mu$ is utilized as the push-forward operator, i.e., $\mathcal{G}_{\mu}(l,\widetilde{\theta}) = \int_{\mathbb{G}^d} \delta(l-\beta( x,\widetilde{\theta})) \mathrm{d}\mu$. For the theoretical properties of a metric, SWD is a true metric that satisfies both symmetry and triangle inequality \cite{bonnotte2013unidimensional}, where the approximation error is obtained and analyzed in \cite{nadjahi2020statistical}. The GSWD defined by \autoref{generalized_SWD} is a true metric if and only if $\beta(\cdot)$ in $\mathcal{G}$ is an injective mapping \cite{chen2021augmented}.

\clearpage
\section{Background on Distributional Representation in Bellman Equation and Temporal Difference Learning}
\label{Section_dist_rep}
\subsection{Reasoning behind Distributional Representation}
The motivation for employing a distributional representation is twofold. Firstly, it provides more comprehensive and richer value-distribution information, thereby enhancing the stability of the learning process. This stability is particularly important for Bayesian learning processes, which often encounter challenges in achieving stable convergence. Secondly, the distributional representation contributes significantly to interpretability. As illustrated in \autoref{TD_convergence_rate_proof1} of the proof, it uses quantiles derived from the distributional representation to formally establish the transparency of the convergence process outlined in \textbf{\cref{theorem_2}}.
\subsection{Distributional Representation in Bellman Equation}
Unlike traditional RL, where the primary objective is to maximize the expected action-value function $Q$, the distributional Bellman equation \cite{bellemare2017distributional} was proposed to approximate and parameterize the entire distribution of future rewards. In the setting of policy evaluation, given a deterministic policy $\pi$, the \textit{Bellman operator} $\mathcal{T}^\pi$ is defined as \cite{bellemare2017distributional,dabney2018distributional}:
\begin{equation}
\mathcal{T}^{\pi} Z(\bm{s},\bm{a})\overset{D}{:=} R(\bm{s},\bm{a})+\gamma Z(S',A')
\label{Bellman_Pred_D_RL}
\end{equation}
where $Z^\pi$ denotes the state-action distribution, and $R(\bm{s},\bm{a})$ denotes the reward distribution. In control setting, a distributional \textit{Bellman optimality operator} $\mathcal{T}$ with quantile approximation is proposed in \cite{dabney2018distributional}:
\begin{equation}
\mathcal{T} Z(\bm{s},\bm{a})\overset{D}{:=} R(\bm{s},\bm{a})+\gamma Z(\bm{s'},{\rm{arg}} \underset{a'}{max}\underset{\bm{p},R}{\mathbb{E}}[Z(\bm{s'},\bm{a'})])
\label{Bellman_Control_D_RL}
\end{equation}
where we let $Z_\theta :=\frac{1}{N}\sum\limits_{i=1}^{N} \delta_{q_i(\bm{s},\bm{a})}$ be a quantile distribution mapping one state-action pair $(\bm{s},\bm{a})$ to a uniform probability distribution supported on $q_i$. Based on \autoref{Bellman_Pred_D_RL}, a contraction is demonstrated \cite{dabney2018distributional} over the Wasserstein metric: $\overset{-}{d}_{\infty}(\Pi_{W_1}\mathcal{T}^{\pi}Z_1,\Pi_{W_1}\mathcal{T}^{\pi}Z_2)\leq \overset{-}{d}_{\infty}(Z_1,Z_2)$, where $\overset{-}{d}_{k}:={\rm{sup}}W_k(Z_1,Z_2)$ denotes the maximal form of the $k$-Wasserstein metrics. $W_k$, $k\in[1,\infty]$ denotes the $k$-Wasserstein distance. $\Pi_{W_1}$ is a quantile approximation under the minimal 1-Wasserstein distance $W_1$.

\subsection{Distributional Representation in Temporal Difference Learning}
Building upon the aforementioned contraction guarantees, we utilize distributional TD learning to estimate the distribution of state-action value, denoted as $Z$. In each iteration, we have the following:
\begin{equation}
\begin{aligned}
\zeta^{i}_{k+1}(\bm{s},\bm{a})&=\zeta^{i}_{k}(\bm{s},\bm{a})+l_{td}\bm{\Delta}^{i}_{k}\\
&=\zeta^{i}_{k}(\bm{s},\bm{a})+l_{td}\times \overset{-}{d}_{\infty}(\Pi_{W_1}\mathcal{T}^{\pi}(h_i(\bm{s},\bm{a},\bm{s}')+\gamma\zeta^{i}_{k}(\bm{s}')),\Pi_{W_1}\mathcal{T}^{\pi}\zeta^{i}_{k}(\bm{s},\bm{a}))
\label{TD_learning}
\end{aligned}
\end{equation}
where $\zeta^{i}_k\in S\times A$ represents the estimated distribution of the state-action distribution $Z$ in the $k$-th TD-learning-iteration for all $i=0,...,p$. The TD learning rate is denoted as $l_{td}$. The function $h_i: S\times A\times S\rightarrow \mathbb{R}$ maps the triple $(\bm{s},\bm{a},\bm{s}')$ to a \textit{real number}. Specifically, $h_i$ is defined as $h_i=r$ when $i=0;$ and $h_i=g_i$ when $i\in[1,n]$. The distributional TD error $\bm{\Delta}^{i}_{k}$ in \autoref{TD_learning} is calculated by $\overset{-}{d}_{\infty}(\Pi_{W_1}\mathcal{T}^{\pi}(h_i(\bm{s},\bm{a},\bm{s}')+\gamma\zeta^{i}_{k}(\bm{s}')),\Pi_{W_1}\mathcal{T}^{\pi}\zeta^{i}_{k}(\bm{s},\bm{a}))$.

In Algorithm~{\autoref{Opt_Rect_Policy_Update}}, the gradient of actor and critic network, denoted as $\delta_{\theta^\mu}$ and $\delta_{\theta^Q}$, can be calculated as follows:
\begin{equation}
\begin{aligned} 
\delta_{\theta^\mu} &= (1/N)\sum\limits{\nabla_{\theta^\mu}}{\pi_{\theta^\mu}}(\bm{s}_n)\mathbb{E}[{\nabla _a}Z_{\theta^Q}(\bm{s}_n,\bm{a})]_{a=\pi_{\theta^\mu}(\bm{s}_n)} \\
\delta_{\theta^Q} &= (1/N)\sum\limits \nabla_{\theta^Q} \overset{-}{d}_{\infty}(\Pi_{W_1}\mathcal{T}^{\pi}Z_{\theta^Q}(\bm{s}_n,\bm{a}_n),\Pi_{W_1}\mathcal{T}^{\pi}\widetilde{g}_{i}(\tau))
\end{aligned} 
\label{grad_AC}
\end{equation}
where $\Pi_{W_1}$ represents a quantile approximation under the minimal 1-Wasserstein distance $W_1$.

\clearpage
\section{Algorithm Details and Proofs}
\label{Section_proof_App}
\subsection{Detailed Derivation of the Objective Function} 
\label{Subsection_object}
The aim of variational inference is to minimize the distance $D(q_\theta(\tau)||p(\tau|\mathcal{O}))={\bf{A-GSWD}}_{k}(q_\theta(\tau),p(\tau|\mathcal{O}))$ between the variational distribution $q_\theta(\tau)$ and the posterior distribution $p(\tau|\mathcal{O})$. Let $\mathcal{P}_{trans}=p\left(\bm{s},\bm{a}|\theta\right)p_D\left(\theta\right)$, $\tilde{x}=x/\mathcal{P}_{trans}$ and $\tilde{y}=y/\mathcal{P}_{trans}$ and recall \textbf{\cref{a_gswd_def}}, i.e., the definition of ${\bf{A-GSWD}}_{k}$. Then, the variational inference can be reformulated to the minimization problem:
\begin{equation}
\begin{aligned} 
&\arg \underset{q_\theta(\tau)}{\rm{min}} {\bf{A-GSWD}}_{k}\left(q_\theta\left(\tau\right),p\left(\tau|\mathcal{O}\right)\right) = \arg \underset{q_\theta(\tau)}{\rm{min}} {\bf{A-GSWD}}_{k}\left(q\left(\bm{a}\right)\cdot\mathcal{P}_{trans},p\left(\mathcal{O}|\tau\right)\cdot\mathcal{P}_{trans}\right)\\ 
&=\arg \underset{q_\theta(\tau)}{\rm{min}} \left(\int_{\mathcal{X}_{\theta}} W_k^{k}\left(\mathcal{A}_{q\left(\bm{a}\right)\cdot\mathcal{P}_{trans}}\left(\cdot,\widetilde{\theta}\right), \mathcal{A} _{p\left(\mathcal{O}|\tau\right)\cdot\mathcal{P}_{trans}}\left(\cdot,\widetilde{\theta}\right)\right) \mathrm{d}\widetilde{\theta}\right)^{\frac{1}{k}} \\ 
&\overset{(\romannumeral1)}{=} \arg \underset{q_\theta(\tau)}{\rm{min}} \left(\int_{\mathcal{X}_{\theta}} W_k^{k}\left(\mathcal{P}_{trans}\cdot\mathcal{A}_{q\left(\bm{a}\right)}\left(\cdot,\widetilde{\theta}\right), \mathcal{P}_{trans}\cdot\mathcal{A} _{p\left(\mathcal{O}|\tau\right)}\left(\cdot,\widetilde{\theta}\right)\right) \mathrm{d}\widetilde{\theta}\right)^{\frac{1}{k}} \\ 
&= \arg \underset{q_\theta(\tau)}{\rm{min}} \inf_{\gamma \in \Gamma(\mathcal{P}_{trans}\cdot\mathcal{A}_{q\left(\bm{a}\right)}, \mathcal{P}_{trans}\cdot\mathcal{A} _{p\left(\mathcal{O}|\tau\right)})}\left(\int_{\mathcal{X}_{\theta}} \int_{\mathcal{X} \times \mathcal{X}} d(x, y)^k \mathrm{d}\gamma(x, y) \mathrm{d}\widetilde{\theta}\right)^{\frac{1}{k}} \\ 
&= \arg \underset{q_\theta(\tau)}{\rm{min}}\inf_{\gamma \in \Gamma(\mathcal{A}_{q\left(\bm{a}\right)}, \mathcal{A} _{p\left(\mathcal{O}|\tau\right)})}\left(\int \int d(\mathcal{P}_{trans}\tilde{x}, \mathcal{P}_{trans}\tilde{y})^k \mathrm{d}\gamma(\mathcal{P}_{trans}\tilde{x}, \mathcal{P}_{trans}\tilde{y}) \mathrm{d}\widetilde{\theta}\right)^{\frac{1}{k}} \\
&\overset{(\romannumeral2)}{=} \arg \underset{q_\theta(\tau)}{\rm{min}}\inf_{\gamma \in \Gamma(\mathcal{A}_{q\left(\bm{a}\right)}, \mathcal{A}_{p\left(\mathcal{O}|\tau\right)})}\left(\int \int \left(\mathcal{P}_{trans}\cdot d(\tilde{x}, \tilde{y})\right)^k \mathrm{d}\gamma(\mathcal{P}_{trans}\tilde{x}, \mathcal{P}_{trans}\tilde{y}) \mathrm{d}\widetilde{\theta}\right)^{\frac{1}{k}} \\
&\overset{(\romannumeral3)}{=}\arg \underset{q_\theta(\tau)}{\rm{min}} \mathcal{P}_{trans}\cdot\inf_{\gamma \in \Gamma(\mathcal{A}_{q\left(\bm{a}\right)}, \mathcal{A}_{p\left(\mathcal{O}|\tau\right)})}\left(\int \int d(\tilde{x}, \tilde{y})^k \mathrm{d}\gamma(\tilde{x}, \tilde{y}) \mathrm{d}\widetilde{\theta}\right)^{\frac{1}{k}} \\
&=\arg \underset{q_\theta(\tau)}{\rm{min}} \mathcal{P}_{trans}\cdot{\bf{A-GSWD}}_{k}\left(q\left(\bm{a}\right),p\left(\mathcal{O}|\tau\right)\right)
\end{aligned}
\label{trans_VI}
\end{equation}
where $(\romannumeral1)$ follows from the push-forward operator definition: $\mathcal{A}_{\mu}(l,\widetilde{\theta}) = \int_{\mathbb{G}^d} \delta(l-\alpha( x,\widetilde{\theta} )) \mathrm{d}\mu$. $(\romannumeral2)$ follows from $d(c\tilde{x},c\tilde{y}) = cd(\tilde{x},\tilde{y}), c\in(0,1)$ as $d$ is a metric. $(\romannumeral3)$ follows from the fact that $\mathrm{d}\gamma(c\tilde{x},c\tilde{y})=\mathrm{d}\gamma(cx,cy)=\mathrm{d}\gamma(x,y)=\mathrm{d}\gamma(\tilde{x},\tilde{y})$, since $\mathrm{d}\gamma(c\tilde{x},c\tilde{y})$ is the measure of the subset of $\mathcal{X}\times\mathcal{X}$, which is just the re-scaled version of the subset $(\tilde{x},\tilde{y})$ by the map $(x,y)\mapsto (x,y)$.

\autoref{trans_VI} presents that the objective can be transformed to the minimization problem, i.e., $\arg \underset{q_\theta(\tau)}{\rm{min}}{\bf{A-GSWD}}_{k}\left(q\left(\bm{a}\right),p\left(\mathcal{O}|\tau\right)\right)$, where $p(\mathcal{O}|\tau)$ represents the optimality likelihood.

\subsection{Definition of Distributional Temporal Difference}
We use distributional TD learning to estimate the distribution of state-action value, denoted as $Z$. In each iteration, we have the following:
\begin{equation}
\begin{aligned}
\zeta^{i}_{k+1}(\bm{s},\bm{a})&=\zeta^{i}_{k}(\bm{s},\bm{a})+l_{td}\bm{\Delta}^{i}_{k}\\
&=\zeta^{i}_{k}(\bm{s},\bm{a})+l_{td}\times \overset{-}{d}_{\infty}(\Pi_{W_1}\mathcal{T}^{\pi}(h_i(\bm{s},\bm{a},\bm{s}')+\gamma\zeta^{i}_{k}(\bm{s}')),\Pi_{W_1}\mathcal{T}^{\pi}\zeta^{i}_{k}(\bm{s},\bm{a}))
\label{TD_learning}
\end{aligned}
\end{equation}
where $\zeta^{i}_k\in S\times A$ represents the estimated distribution of the state-action distribution $Z$ in the $k$-th TD-learning-iteration for all $i=0,...,p$. The TD learning rate is denoted as $l_{td}$. The function $h_i: S\times A\times S\rightarrow \mathbb{R}$ maps the triple $(\bm{s},\bm{a},\bm{s}')$ to a \textit{real number}. Specifically, $h_i$ is defined as $h_i=r$ when $i=0;$ and $h_i=g^i$ when $i\in[1,p]$. The distributional TD error $\bm{\Delta}^{i}_{k}$ in \autoref{TD_learning} is calculated by $\overset{-}{d}_{\infty}(\Pi_{W_1}\mathcal{T}^{\pi}(h_i(\bm{s},\bm{a},\bm{s}')+\gamma\zeta^{i}_{k}(\bm{s}')),\Pi_{W_1}\mathcal{T}^{\pi}\zeta^{i}_{k}(\bm{s},\bm{a}))$.

\subsection{Proofs}
\label{Section_proof}
Here we present the proofs of \textbf{\cref{proposition_1}} \textit{(Pseudo-metric)}, \textbf{\cref{theorem_1}} \textit{(Global Convergence)} and \textbf{\cref{theorem_2}} \textit{(Global Convergence Rate)}, as outlined in \textbf{Section \ref{Section_formal}}. The associated propositions, namely \textbf{\cref{proposition_2}} \textit{(Policy Evaluation)} and \textbf{\cref{proposition_3}} \textit{(Policy Improvement)}, are explicated and clarified by their respective proofs below.

\textbf{\cref{proposition_1}.} \textit{(Pseudo-metric)}\textit{: Given two probability measures $\mu,\nu\in P_k(\mathcal{X})$ and a mapping $\alpha: \mathcal{X}\rightarrow\mathcal{R}_{\widetilde{\theta}}$, the adaptive slicing A-GSWD, defined in \textbf{\cref{a_gswd_def}}, with order $k$ in the range $[1,\infty)$, is a pseudo-metric. This pseudo-metric satisfies non-negativity, symmetry, the triangle inequality, and ${\bf{A-GSWD}}_{k}(\mu,\mu)=0$.}

\textit{Proof}:
The non-negativity property naturally arises from the fact that the Wasserstein distance $W_k$ is a metric \cite{villani2009optimal}. To prove symmetry, since the k-Wasserstein distance is a metric~\cite{villani2009optimal}: 
\begin{equation}
W_k\left(\mathcal{A}_\mu(\cdot,\widetilde{\theta};\alpha),\mathcal{A}_\nu(\cdot,\widetilde{\theta})\right)=W_k\left(\mathcal{A}_\nu(\cdot,\widetilde{\theta}),\mathcal{A}_\mu(\cdot,\widetilde{\theta})\right)\nonumber
\end{equation}

Thus, there exists \cite{chen2021augmented}: 
\begin{equation}
\begin{aligned}
\bf{A-GSWD}_{k}(\mu,\nu)&=\left(\int_{\mathcal{R}_{\widetilde{\theta}}} W_k^k\left(\mathcal{A}_\mu(\cdot,\widetilde{\theta}),\mathcal{A}_\nu(\cdot,\widetilde{\theta})\right)\mathrm{d}\widetilde{\theta}\right)^{\frac{1}{k}}\\
&=\left(\int_{\mathcal{R}_{\widetilde{\theta}}} W_k^k\left(\mathcal{A}_\nu(\cdot,\widetilde{\theta}),\mathcal{A}_\mu(\cdot,\widetilde{\theta})\right)\mathrm{d}\widetilde{\theta}\right)^{\frac{1}{k}}=\bf{A-GSWD}_{k}(\nu,\mu)\nonumber
\end{aligned}
\end{equation}

Therefore, symmetry holds. Then, we prove the triangle inequality. Since the triangle inequality holds for the Wasserstein distance, we can obtain $W_k\left(\mathcal{A}_{\mu_1}, \mathcal{A} _{\mu_3}\right)\leq W_k\left(\mathcal{A}_{\mu_1}, \mathcal{A} _{\mu_2}\right)+W_k\left(\mathcal{A}_{\mu_2}, \mathcal{A} _{\mu_3}\right)$. Thus, there exists:
\begin{equation}
\begin{aligned}
&{\bf{A-GSWD}}_{k}(\mu_1,\mu_3) = \left(\int_{\mathcal{R}_{\widetilde{\theta}}} W_k^{k}\left(\mathcal{A}_{\mu_1}, \mathcal{A} _{\mu_3}\right) \mathrm{d}\widetilde{\theta}\right)^{\frac{1}{k}}\\
&\leq \left(\int_{\mathcal{R}_{\widetilde{\theta}}} W_k^{k}\left(\mathcal{A}_{\mu_1}, \mathcal{A} _{\mu_2}\right)+W_k^{k}\left(\mathcal{A}_{\mu_2}, \mathcal{A} _{\mu_3}\right) \mathrm{d}\widetilde{\theta}\right)^{\frac{1}{k}}\\
&\leq \left(\int_{\mathcal{R}_{\widetilde{\theta}}} W_k^{k}\left(\mathcal{A}_{\mu_1}, \mathcal{A} _{\mu_2}\right) \mathrm{d}\widetilde{\theta}\right)^{\frac{1}{k}} +\left(\int_{\mathcal{R}_{\widetilde{\theta}}}W_k^{k}\left(\mathcal{A}_{\mu_2}, \mathcal{A} _{\mu_3}\right) \mathrm{d}\widetilde{\theta}\right)^{\frac{1}{k}}
\end{aligned}
\label{A_GSWD_in}
\end{equation}
where the derivation of \autoref{A_GSWD_in} is based on the Minkowski inequality \cite{bahouri2011fourier}, which establishes that ${\bf{A-GSWD}}_{k}$ satisfies the triangle inequality. Since $W_k(\mu,\mu)=0$ for any $\mu$~\cite{villani2009optimal}, we have ${\bf{A-GSWD}}_{k}(\mu,\mu)=\left(\int_{\mathcal{R}_{\widetilde{\theta}}} W_k^{k}\left(\mathcal{A}_{\mu}, \mathcal{A} _{\mu}\right) \mathrm{d}\widetilde{\theta}\right)^{\frac{1}{k}}=0$.

Therefore, A-GSWD is a pseudo-metric that satisfies non-negativity, symmetry, the triangle inequality, and ${\bf{A-GSWD}}_{k}(\mu,\mu)=0$. \hfill $\blacksquare$

\textit{\textbf{\cref{remark}.} The adaptive slicing A-GSWD, with order $k\in[1,\infty)$, is a true metric \textit{if and only if} the AGRT $\mathcal{A}$, defined in \textbf{\cref{a_gswd_def}}, is an injective mapping.}

\textit{Proof}:
Given the indiscernibility property for the Wasserstein distance $W_k$ \cite{villani2009optimal}, it follows $W_k(\mu_1, \mu_2)=0 \; \text{if and only if} \; \mu_1=\mu_2$. Consequently, ${\bf{A-GSWD}}_{k}(\mu_1, \mu_2)=0\quad  \text{is equivalent to} \quad \mathcal{A}_{\mu_1}(l,\widetilde{\theta})=\mathcal{A}_{\mu_2}(l,\widetilde{\theta})$. The equality $\mathcal{A}_{\mu_1}(l,\widetilde{\theta})=\mathcal{A}_{\mu_2}(l,\widetilde{\theta})$ implies $\mu_1=\mu_2$ if and only if $\mathcal{A}$ is an injective mapping.

Therefore, A-GSWD is a metric if and only if $\mathcal{A}_{\mu_1}(l,\widetilde{\theta})=\mathcal{A}_{\mu_2}(l,\widetilde{\theta})$ implies $\mu_1=\mu_2$, i.e., the AGRT $\mathcal{A}$ is an injective mapping. \hfill $\blacksquare$

\begin{proposition}
\label{proposition_2}
\textit{(Policy Evaluation) \cite{dabney2018distributional,wang2023quadue})}: we consider a quantile approximation $\Pi_{W_1}$ under the minimal 1-Wasserstein distance $W_1$, the \textit{Bellman operator} $\mathcal{T}^\pi$ under a deterministic policy $\pi$ and $Z_{k+1}(\bm{s},\bm{a})=\Pi_{W_1}\mathcal{T}^{\pi} Z_k(\bm{s},\bm{a})$. The sequence $Z_{k}(\bm{s},\bm{a})$ converges to a unique fixed point $\overset{\sim}{Z_\pi}$ under the maximal form of $\infty$-Wasserstein metric $\overset{-}{d}_{\infty}$.
\end{proposition} 

\textit{Proof}:
We recall a contraction proved in \cite{dabney2018distributional} over the Wasserstein Metric:
\begin{equation}
\overset{-}{d}_{\infty}(\Pi_{W_1}\mathcal{T}^{\pi}Z_1,\Pi_{W_1}\mathcal{T}^{\pi}Z_2)\leq \overset{-}{d}_{\infty}(Z_1,Z_2)
\label{contraction_WM}
\end{equation}
where \autoref{contraction_WM} implies that the combined operator $\Pi_{W_1} \mathcal{T}^\pi$ is an $\infty$-contraction. Based on Banach’s fixed point theorem, $\mathcal{T}^\pi$ has a unique fixed point, i.e., $\overset{\sim}{Z_\pi}$. Furthermore, the definition of \textit{Bellman optimality operator}, defined as \autoref{Bellman_Control_D_RL}, implies that all moments of $Z$ are bounded. Therefore, we conclude that the sequence $Z_{k}(\bm{s},\bm{a})$ converges to $\overset{\sim}{Z_\pi}$ in $\overset{-}{d}_{\infty}$ for $p\in[1,\infty]$. \hfill $\blacksquare$ 

\begin{proposition}
\label{proposition_3}
\textit{(Policy Improvement)}: Given an old policy $\bm{\pi}_{\bm{old}}$, a new policy $\bm{\pi}_{\bm{new}}$ and $Q(s, a)=\mathbb{E}[Z(s,a)]$, suppose \cref{assumption_1} holds, there exists $Q^{\bm{\pi}_{\bm{new}}}(s, a) \geq Q^{\bm{\pi}_{\bm{old}}}(s, a)$ when performing Algorithm~{\autoref{Opt_Rect_Policy_Update}}, $\forall s\in \mathcal{S}$ and $\forall a\in \mathcal{A}$ \textit{if and only if} the reward operator family $\mathcal{F}=\left\{\mathcal{F}_r,\mathcal{F}_g\right\}$ satisfies the both \textbf{\cref{conditions}}.
\end{proposition} 


\textit{Proof}:
We recall that $\left\{\mathcal{F}_r,\mathcal{F}_g\right\}$ are two operators defined as $\widetilde{r}(\tau):=\mathcal{F}_r\cdot p\left(\mathcal{O}_r|\tau\right)$ and $\widetilde{g}^{i}(\tau):=\mathcal{F}_g\cdot p\left(\mathcal{O}_{g_i}|\tau\right)$, respectively. Suppose \cref{assumption_1} holds. Since the two optimization objectives in policy updating, i.e., ${\rm{max}}\mathbb{E}[\mathcal{F}_r\cdot p\left(\mathcal{O}_r|\tau\right)]$ and ${\rm{min}}\mathbb{E}[\mathcal{F}_g\cdot p\left(\mathcal{O}_{g_i}|\tau\right)]$ (see \textbf{Section \ref{Subsec_orpodr}}), and $p\left(\mathcal{O}|\tau\right)$ is defined on $(0,1]$, we can conclude the both \textbf{\cref{conditions}} that (\textit{i}) $\mathcal{F}_r$ is monotonically increasing and continuously defined on $(0,1]$, and the range covers $[r_{\min}, r_{\max}]$;
(\textit{ii}) $\mathcal{F}_g$ is monotonically decreasing and continuously defined on $(0,1]$, and the range covers $[r_{\min}, r_{\max}]$.

Then based on \autoref{Bellman_Control_D_RL}, there exists:
\begin{equation}
\begin{aligned}
V^\pi(s_{t})&=\mathbb{E}[{Q(s_{t},\pi(s_{t}))}]\leq \underset{a'\in \mathcal{A}}{\rm{max}} \mathbb{E}[{Q(s_{t},a')}] \\
&=\mathbb{E}[{Q(s_{t},{\pi'}(s_{t}))}]
\end{aligned}
\label{Q_V}
\end{equation}
where $\mathbb{E}_\pi[\cdot]=\sum_{a\in A}\bm{\pi}(a|s)[\cdot]$, and $V^\pi(s)=\mathbb{E}_\pi \mathbb{E}[Z_k(s,a)]$ is the value function. According to \autoref{Q_V} and \autoref{Bellman_Control_D_RL}, it yields:
\begin{equation}
\begin{aligned}
Q^{\bm{\pi}_{\bm{old}}} &=Q^{\bm{\pi}_{\bm{old}}}(s_{t},\bm{\pi}_{\bm{old}}(s_{t})) \\
&= r_{t+1}+\gamma \mathbb{E}_{s_{t+1}} \mathbb{E}_{\bm{\pi}_{\bm{old}}} Q^{\bm{\pi}_{\bm{old}}}(s_{t+1},{\bm{\pi}_{\bm{old}}}(s_{t+1}))\\
&\overset{(i)}{\leq} r_{t+1}+\gamma \mathbb{E}_{s_{t+1}} \mathbb{E}_{\bm{\pi}_{\bm{new}}}{Q^{\bm{\pi}_{\bm{old}}}(s_{t+1},{\bm{\pi}_{\bm{new}}}(s_{t+1}))}\\
&\leq r_{t+1}+\mathbb{E}_{s_{t+1}} \mathbb{E}_{\bm{\pi}_{\bm{new}}} [\gamma r_{t+2} \\ &+ {\gamma^2} \mathbb{E}_{s_{t+2}}{Q^{\bm{\pi}_{\bm{old}}}(s_{t+2},{\bm{\pi}_{\bm{new}}}(s_{t+2}))}|]\\
&\leq r_{t+1}+\mathbb{E}_{s_{t+1}} \mathbb{E}_{\bm{\pi}_{\bm{new}}} [\gamma r_{t+2} + {\gamma^2}r_{t+3} + ...]\\
&= r_{t+1}+\mathbb{E}_{s_{t+1}} V^{\bm{\pi}_{\bm{new}}}(s_{t+1})\\
&=Q^{\bm{\pi}_{\bm{new}}}
\end{aligned}
\label{policy_improvement}
\end{equation} 

where $(i)$ relies on \autoref{Q_V}, and $\bm{\pi}_{\bm{new}}$ corresponds to the maximum $Q$ in the Bellman function. Therefore, we have $Q^{\bm{\pi}_{\bm{new}}}(s, a) \geq Q^{\bm{\pi}_{\bm{old}}}(s, a)$. \hfill $\blacksquare$

Then we provide \textbf{\cref{lemma_1}} and the proof of \textbf{\cref{theorem_1}}.

\begin{lemma}
\label{lemma_1}
\cite{bellemare2017distributional}: The \textit{Bellman operator} $\mathcal{T}^\pi$ is a $p$-contraction under the $p$-Wasserstein metric $\overset{-}{d}_{p}$.
\end{lemma}

\textit{\textbf{\cref{theorem_1}.} \textit{(Global Convergence)}: Given the policy in the $i$-th policy improvement $\bm{\pi}^{\bm{i}}$, $\bm{\pi}^{\bm{i}} \rightarrow \bm{\pi}^{*}$ and $i\rightarrow\infty$, suppose \cref{assumption_1} holds,} there exists $Q^{\bm  {\pi}^{*}}(s, a) \geq Q^{\bm{\pi}^{\bm{i}}}(s, a)$ \textit{if and only if} the reward operator family $\mathcal{F}$ satisfies the both \textbf{\cref{conditions}}.

\textit{Proof}:
Since \textbf{\cref{proposition_3}} suggests $Q^{\bm{\pi}_{\bm{i+1}}}(s, a) \geq Q^{\bm{\pi}_{\bm{i}}}(s, a)$, the sequence $Q^{\bm{\pi}_{\bm{i}}}(s, a)$ is monotonically increasing \textit{if and only if} the reward operator family $\mathcal{F}$ satisfies the both \textbf{\cref{conditions}}. Furthermore, \textbf{\cref{lemma_1}} implies that the the state-action distribution $Z$ over $\mathbb{R}$ has bounded $p$-th moment, so the first moment of $Z$, i.e., $Q^{\bm{\pi}_{\bm{i}}}(s, a)$, is upper bounded. Therefore, the sequence $Q^{\bm{\pi}_{\bm{i}}}(s, a)$ converges to an upper limit $Q^{\bm{\pi}_{*}}(s, a)$ with $\forall s\in \mathcal{S}$ and $\forall a\in \mathcal{A}$. \hfill $\blacksquare$

To prove \textbf{\cref{theorem_2}}, we provide \textbf{\cref{lemma_2}} and its proof below.

\begin{lemma}
\label{lemma_2}
\textit{(Convergence rate of neural TD learning)}: Let $m$ be the width of the actor-critic networks, and $\overset{-}{Z_{t}}=\frac{1}{N}\sum\limits_{i=1}^{N} \delta_{q_i(\bm{s},\bm{a})}$ be an estimator of $Z^{i}_{t}$. Suppose \cref{assumption_2} holds, In the TD learning, with probability at least $1-\delta$, there exists
\begin{equation}
\left \| \Pi_{W_1}\overset{-}{Z_{t}}-\Pi_{W_1}{Z_{t}^*} \right \| \leq \Theta(m^{-\frac{H}{4}}) +\Theta({[(1-\gamma)K]^{-\frac{1}{2}}}[1+\log^{\frac{1}{2}}\delta^{-1}])
\label{convergence_rate_in_TD}
\end{equation}
\end{lemma}

\textit{Proof}:
Utilizing Gluing Lemma~\cite{villani2009wasserstein,clement2008elementary} for Wasserstein distance $W_p$, where we employ the $1$-Wasserstein distance $W_1$ and the one-dimensional quantile $q_t^{i,*}$, we establish:
\begin{equation}
\begin{aligned}
&\left \| \Pi_{W_1}\overset{-}{Z_{t}}-\Pi_{W_1}{Z_{t}^*} \right \|
{=}\sum\limits_{i=1}^{N}\left \| \overset{-}{q_t^i}-q_t^{i,*} \right \| \\
&=\sum\limits_{i=1}^{N}\left \| f_i^{(H)}((\bm{s},\bm{a}),\theta_{K_{td}}^Q)-f_i^{(H)}((\bm{s},\bm{a}),\theta^{Q^*})\right \| \\
&\leq \sum\limits_{i=1}^{N}\left \| f_i^{(H)}((\bm{s},\bm{a}),\theta_{K_{td}}^Q)-f_{0,i}^{(H)}((\bm{s},\bm{a}),\theta^Q)\right \| \\
&+ \sum\limits_{i=1}^{N} \left \| f_{0,i}^{(H)}((\bm{s},\bm{a}),\theta_{K_{td}}^Q)-f_i^{(H)}((\bm{s},\bm{a}),\theta^{Q^*})\right \| \\
&\overset{(i)}{\leq}\Theta(m^{-\frac{H}{4}})+\sum\limits_{i=1}^{N} \left \| f_{0,i}^{(H)}((\bm{s},\bm{a}),\theta_{K_{td}}^Q)-f_i^{(H)}((\bm{s},\bm{a}),\theta^{Q^*})\right \| \\
&\overset{(ii)}{\leq}\Theta(m^{-\frac{H}{4}})+\Theta({[(1-\gamma)K_{td}]^{-\frac{1}{2}}}[1+\log^{\frac{1}{2}}\delta^{-1}])
\end{aligned}
\label{TD_convergence_rate_proof1}
\end{equation}
where  $H$ denotes the layers of the neural network. Suppose \cref{assumption_2} holds. Then (i) follows from \textbf{Lemma 5.1} in \cite{cai2019neural}, where each quantile represents as a form of local linearization \cite{koenker2005quantile,gannoun2007comparison}: 
\begin{equation}
\begin{aligned}
&\sum\limits_{i=1}^{N} \left \| f_i^{(H)}((\bm{s},\bm{a}),\theta_{K_{td}}^Q)-f_{0,i}^{(H)}((\bm{s},\bm{a}),\theta^Q)\right \|^2 \leq \frac{1}{m^{H}}\sum\limits_{i=1}^{N} b_r\\
&\left |[(\mathbf{1}(W_i^{(h)} x_i^{(h-1)}>0)-\mathbf{1}(W_i^{(0)} x_i^{(h-1)}>0))\cdot W_i^{(h)} x_i^{(h-1)}]^2\right | \\
&\leq \frac{4C_0}{m^{H}} \sum\limits_{i=1}^{N}[\sum\limits_{r=1}^{m}\mathbf{1}(\left | W_{i,r}^{(0)} x_i^{(h-1)}\right |\leq\left \| W_{i,r}^{(h)}-W_{i,r}^{(0)}\right \|_2 )] \\
&\leq \frac{4C_0}{m^{H}} (\sum\limits_{r=1}^{m}\left \|W_{i,r}^{(h)}-W_{i,r}^{(0)}\right \|_2^2)^{\frac{1}{2}}(\sum\limits_{r=1}^{m}\left \|\frac{1}{W_{i,r}^{(0)}}\right \|_2^2)^{\frac{1}{2}}\leq \frac{4C_0C_1}{m^{\frac{H}{2}}}
\end{aligned}
\label{TD_convergence_rate_proof2}
\end{equation}
where the constant $C_0>0$ and $C_1>0$. Thus we upper bound $\sum\limits_{i=1}^{N} \left \| f_i^{(H)}-f_{0,i}^{(H)}\right \| \leq \Theta(m^{-\frac{H}{4}})$, which holds (i) in \autoref{TD_convergence_rate_proof1}. Suppose \cref{assumption_3} holds. Then (ii) follows from \textbf{Lemma 1} in \cite{rahimi2008weighted}, with probability at least $1-\delta$, there exists: 
\begin{equation}
\begin{aligned}
&\sum\limits_{i=1}^{N} \left \| f_{0,i}^{(H)}((\bm{s},\bm{a}),\theta_{K_{td}}^Q)-f_i^{(H)}((\bm{s},\bm{a}),\theta^{Q^*})\right \| \\
&\leq \frac{1}{\sqrt{1-\gamma}}\sum\limits_{i=1}^{N} \left \| f_{0,i}^{(H)}((\bm{s},\bm{a}),\theta_{K_{td}}^{Q_\pi})-f_i^{(H)}((\bm{s},\bm{a}),\theta^{Q^*})\right \| \\
&\leq \frac{C_3}{\sqrt{(1-\gamma)K_{td}}}(1+\sqrt{\log{\frac{1}{\delta}}})
\end{aligned}
\label{TD_convergence_rate_proof3}
\end{equation}
where (iii) holds, and therefore \autoref{TD_convergence_rate_proof1} holds. \hfill $\blacksquare$

\textit{\textbf{\cref{theorem_2}.} \textit{(Global Convergence Rate)}: Let $m$ and $H$ be the width and the layer of the neural network, $K_{td}=(1-\gamma)^{-\frac{3}{2}}m^{\frac{H}{2}}$ be the iterations required for convergence of the distributional TD learning (defined in \autoref{TD_learning}), $l_{Q}=\frac{1}{\sqrt{T}}$ be the policy update (in \textit{Line 4} of Algorithm~{\autoref{Opt_Rect_Policy_Update}}) and  $\bm{\tau}_{c}=\Theta(\frac{1}{(1-\gamma)\sqrt{T}})+\Theta(\frac{1}{(1-\gamma)Tm^{\frac{H}{4}}})$ be the tolerance (in \textit{Line 3} of Algorithm~{\autoref{Opt_Rect_Policy_Update}}). Suppose  Assumptions 1-3 hold.} There exists a global convergence rate of $\Theta(1/\sqrt{T})$, and a sublinear rate of $\Theta(1/\sqrt{T})$ if the constraints are violated with an error of $\Theta(1/{m^{\frac{H}{4}}})$, with probability at least $1-\delta$. Importantly, this conclusion holds if and only if the reward operator family $\mathcal{F}$ satisfies both \textbf{\cref{conditions}}.

\textit{Proof}:
\textbf{\cref{proposition_3}} suggests that the sequence $Q^{\bm{\pi}_{\bm{i}}}(s, a)$ achieves global convergence \textit{if and only if} the reward operator family $\mathcal{F}$ satisfies \textbf{\cref{conditions}}. Then we let $\triangle_{\theta^Q}=\theta_{t+1}^Q-\theta_{t}^Q$, and suppose the critic networks are $H$-layer neural networks. Based on \textbf{Lemma 6.1} in \cite{kakade2002approximately}, there exists
\begin{equation}
\begin{aligned}
&(1-\gamma)[\mathcal{J}_r(\bm{\pi^*})-{\mathcal{J}_r}(\bm{\pi}_t)] \\
&=\mathbb{E}[Q_{\pi_t}(\bm{s},\bm{a})-\mathbb{E}Q_{\pi_t}(\bm{s},\bm{a}')] \\
&=\mathbb{E}[\nabla_{\theta}f^{(H)}((\bm{s},\bm{a}),\theta^Q)^{\mathrm{T}}-\mathbb{E}[\nabla_{\theta}f^{(H)}((\bm{s},\bm{a}'),\theta^Q)^{\mathrm{T}}]]\triangle_{\theta^Q}\\
&\ +\mathbb{E}[Q_{\pi_t}(\bm{s},\bm{a})-\nabla_{\theta}f^{(H)}((\bm{s},\bm{a}),\theta^Q)^{\mathrm{T}}\triangle_{\theta^Q}]\\
&\ +\mathbb{E}[\nabla_{\theta}f^{(H)}((\bm{s},\bm{a}'),\theta^Q)^{\mathrm{T}}\triangle_{\theta^Q}-Q_{\pi_t}(\bm{s},\bm{a}')]\\
&=\frac{1}{l_Q}\big[l_Q \mathbb{E}[\nabla_{\theta} \log(\bm{\pi}_t(\bm{a}|\bm{s}))^{\mathrm{T}}]\triangle_{\theta^Q}-\frac{l_{Q}^2\mathcal{L}_f}{2}\left \|\triangle_{\theta^Q}\right \|_2^2 \big]\\
&\ +\mathbb{E}[Q_{\pi_t}(\bm{s},\bm{a})-\nabla_{\theta}f^{(H)}((\bm{s},\bm{a}),\theta^Q)^{\mathrm{T}}\triangle_{\theta^Q}]+\frac{l_{Q}\mathcal{L}_f}{2}\left \|\triangle_{\theta^Q}\right \|_2^2\\
&\ +\mathbb{E}[\nabla_{\theta}f^{(H)}((\bm{s},\bm{a}'),\theta^Q)^{\mathrm{T}}\triangle_{\theta^Q}-Q_{\pi_t}(\bm{s},\bm{a}')]\\
&\overset{(i)}{\leq}\frac{1}{l_Q} \mathbb{E}[\log(\frac{\bm{\pi}_{t+1}(\bm{a}|\bm{s})}{\bm{\pi}_t(\bm{a}|\bm{s})})]+\frac{l_{Q}\mathcal{L}_f}{2}\left \|\triangle_{\theta^Q}\right \|_2^2\\
&\ +\sqrt{\mathbb{E}[Q_{\pi_t}(\bm{s},\bm{a})-f^{(H)}((\bm{s},\bm{a}),\triangle_{\theta^Q})]^2}\\
&\ +\sqrt{\mathbb{E}[f^{(H)}((\bm{s},\bm{a}),\triangle_{\theta^Q})-\nabla_{\theta}f^{(H)}((\bm{s},\bm{a}),\theta^Q)^{\mathrm{T}}\triangle_{\theta^Q}]^2}\\
&\ +\sqrt{\mathbb{E}[\nabla_{\theta}f^{(H)}((\bm{s},\bm{a}'),\theta^Q)^{\mathrm{T}}\triangle_{\theta^Q}-f^{(H)}((\bm{s},\bm{a}'),\triangle_{\theta^Q})]^2} \\
&\ +\sqrt{\mathbb{E}[f^{(H)}((\bm{s},\bm{a}'),\triangle_{\theta^Q})-Q_{\pi_t}(\bm{s},\bm{a}')]^2}\\
&\ =\frac{1}{l_Q}\big[\mathbb{E}[\mathcal{D}_{KL}(\bm{\pi^*}||\bm{\pi}_{t})]-\mathbb{E}[\mathcal{D}_{KL}(\bm{\pi^*}||\bm{\pi}_{t+1})]\big]\\
&\ +2\sqrt{\mathbb{E}[f^{(H)}((\bm{s},\bm{a}),\triangle_{\theta^Q})-\nabla_{\theta}f^{(H)}((\bm{s},\bm{a}),\theta^Q)^{\mathrm{T}}\triangle_{\theta^Q}]^2}\\
&\ +2\sqrt{\mathbb{E}[Q_{\pi_t}(\bm{s},\bm{a})-f^{(H)}((\bm{s},\bm{a}),\triangle_{\theta^Q})]^2}+\frac{l_{Q}\mathcal{L}_f}{2}\left \|\triangle_{\theta^Q}\right \|_2^2
\end{aligned}
\label{Global_convergence_proof1}
\end{equation}
where (i) follows from the $\mathcal{L}_f$-Lipschitz property of $\log(\bm{\pi}_t(\bm{a}|\bm{s}))$. Suppose \cref{assumption_2} holds. Next, we upper bound the term $\sqrt{\mathbb{E}[f^{(H)}((\bm{s},\bm{a}),\triangle_{\theta^Q})-\nabla_{\theta}f^{(H)}((\bm{s},\bm{a}),\theta^Q)^{\mathrm{T}}\triangle_{\theta^Q}]^2}$ as shown below.
\begin{equation}
\begin{aligned}
&\sqrt{\mathbb{E}[f^{(H)}((\bm{s},\bm{a}),\triangle_{\theta^Q})-\nabla_{\theta}f^{(H)}((\bm{s},\bm{a}),\theta^Q)^{\mathrm{T}}\triangle_{\theta^Q}]^2}\\
&=\sum\limits_{i=1}^{N}\left \| f_i^{(H)}((\bm{s},\bm{a}),\triangle_{\theta^Q})-\nabla_{\theta}f_i^{(H)}((\bm{s},\bm{a}),\theta^Q)^{\mathrm{T}}\triangle_{\theta^Q}\right \| \\
&\leq \sum\limits_{i=1}^{N}\big[ \left \| f_i^{(H)}((\bm{s},\bm{a}),\triangle_{\theta^Q})-\nabla_{\theta}f_{0,i}^{(H)}((\bm{s},\bm{a}),\theta^Q)^{\mathrm{T}}\triangle_{\theta^Q} \right \| \\
&+\left \|\nabla_{\theta}f_{0,i}^{(H)}((\bm{s},\bm{a}),\theta^Q)^{\mathrm{T}}\triangle_{\theta^Q}-\nabla_{\theta}f_i^{(H)}((\bm{s},\bm{a}),\theta^Q)^{\mathrm{T}}\triangle_{\theta^Q}\right \| \big]\\
&=2\sum\limits_{i=1}^{N}\left \| f_i^{(H)}((\bm{s},\bm{a}),\triangle_{\theta^Q})-f_{0,i}^{(H)}((\bm{s},\bm{a}),\triangle_{\theta^Q})\right \| \\
&\overset{(ii)}{\leq} \frac{4\sqrt{C_0C_1}}{m^{\frac{H}{4}}}
\end{aligned}
\label{Global_convergence_proof2}
\end{equation}
where (ii) follows from \autoref{TD_convergence_rate_proof2}. Suppose \cref{assumption_3} holds. Then, in order to upper bound $\sqrt{\mathbb{E}[Q_{\pi_t}(\bm{s},\bm{a})-f^{(H)}((\bm{s},\bm{a}),\triangle_{\theta^Q})]^2}$, taking expectation of \autoref{Global_convergence_proof1} from $t=0$ to $T-1$, yields
\begin{equation}
\begin{aligned}
&(1-\gamma)\big[\mathcal{J}_r(\bm{\pi^*})-\mathbb{E}[{\mathcal{J}_r}(\bm{\pi})]\big]\\
&=(1-\gamma)\frac{1}{T}\sum\limits_{t=0}^{T-1}[\mathcal{J}_r(\bm{\pi^*})-{\mathcal{J}_r}(\bm{\pi}_t)]\\
&\leq \frac{1}{T}\big[\frac{1}{l_Q}\mathbb{E}[\mathcal{D}_{KL}(\bm{\pi^*}||\bm{\pi}_{t})]+\frac{8T\sqrt{C_0C_1}}{m^{\frac{H}{4}}}+\frac{Tl_{Q}\mathcal{L}_f}{2}d_{\theta}^2\\
&+2\sum\limits_{t=0}^{T-1}\sum\limits_{i=1}^{N}\left \| f_i^{(H)}((\bm{s},\bm{a}),\theta_{t+1}^Q-\theta_{t}^Q)-f_i^{(H)}((\bm{s},\bm{a}),\theta^{Q^*})\right \|\big] \\
&= \frac{\mathbb{E}[\mathcal{D}_{KL}(\bm{\pi^*}||\bm{\pi}_{t})]}{l_QT}+\frac{8\sqrt{C_0C_1}}{m^{\frac{H}{4}}}+\frac{l_{Q}\mathcal{L}_f}{2}d_{\theta}^2\\
&\ +\frac{2}{T}\sum\limits_{i=1}^{N}\left \| f_i^{(H)}((\bm{s},\bm{a}),\theta_{K_{td},t}^Q)-f_i^{(H)}((\bm{s},\bm{a}),\theta^{Q^*})\right \| \\
&\overset{(iii)}{\leq} \frac{\mathbb{E}[\mathcal{D}_{KL}(\bm{\pi^*}||\bm{\pi}_{t})]}{l_QT}+\frac{8\sqrt{C_0C_1}}{m^{\frac{H}{4}}}+\frac{l_{Q}\mathcal{L}_f}{2}d_{\theta}^2\\
&\ +\frac{4\sqrt{C_0C_1}}{Tm^{\frac{H}{4}}}+\frac{2C_3}{T\sqrt{(1-\gamma)K_{td}}}(1+\sqrt{\log{\frac{1}{\delta}}})
\end{aligned}
\label{Global_convergence_proof3}
\end{equation}
where (iii) follows from \textbf{\cref{lemma_2}} (\autoref{TD_convergence_rate_proof1}). Thus, substituting $K_{td}=(1-\gamma)^{-1}m^{\frac{H}{2}}$ and $l_{Q}=\Theta(1/\sqrt{T})$ into \autoref{Global_convergence_proof3}, with probability at least $1-\delta$, yields: 
\begin{equation}
\begin{aligned}
&\mathcal{J}_r(\bm{\pi^*})-\mathbb{E}[{\mathcal{J}_r}(\bm{\pi})]\leq C_5\frac{1}{(1-\gamma)\sqrt{T}}+C_6\frac{1}{(1-\gamma)m^{\frac{H}{4}}}\\
&\ +C_7\frac{1}{(1-\gamma)Tm^{\frac{H}{4}}}+2C_3\frac{\sqrt{\log{\frac{1}{\delta}}}}{(1-\gamma)Tm^{\frac{H}{4}}}\\
&\leq\Theta(\frac{1}{(1-\gamma)\sqrt{T}})+\Theta(\frac{1}{(1-\gamma)Tm^{\frac{H}{4}}}\sqrt{\log{\frac{1}{\delta}}})
\end{aligned}
\label{Global_convergence_proof4}
\end{equation}
where $C_5=\mathbb{E}[\mathcal{D}_{KL}(\bm{\pi^*}||\bm{\pi}_{t})]+\frac{\mathcal{L}_fd_{\theta}^2}{2}$, $C_6=8\sqrt{C_0C_1}$ and $C_7=4\sqrt{C_0C_1}+2C_3$. Therefore, there exists:
\begin{equation}
\begin{aligned}
\mathcal{J}_r(\bm{\pi^*})-\mathbb{E}[{\mathcal{J}_r}(&\bm{\pi})]\leq\Theta(\frac{1}{(1-\gamma)\sqrt{T}})\\
&+\Theta(\frac{1}{(1-\gamma)Tm^{\frac{H}{4}}}\sqrt{\log{\frac{1}{\delta}}})
\end{aligned}
\label{Global_convergence1}
\end{equation}
where \autoref{Global_convergence1} suggests that there exists a global convergence rate of $\Theta(1/\sqrt{T})$, with probability at least $1-\delta$.

Following \textit{Line 6} in Algorithm~\autoref{Opt_Rect_Policy_Update} and recalling \autoref{Global_convergence_proof1}, \autoref{Global_convergence_proof2} and \autoref{Global_convergence_proof3}, the convergence process is similarly stated for the constraint approximation ${\mathcal{J}^{i}_g}(\bm{\pi}),\ \forall i\in[1,p]$ here
\begin{equation}
\begin{aligned}
\mathbb{E}[{\mathcal{J}^i_g}(\bm{\pi})]-\mathcal{J}^i_g(\bm{\pi^*})&\leq\Theta(\frac{1}{(1-\gamma)\sqrt{T}})\\
&\ +\Theta(\frac{1}{(1-\gamma)Tm^{\frac{H}{4}}}\sqrt{\log{\frac{1}{\delta}}})
\end{aligned}
\label{Global_convergence_proof5}
\end{equation}
the constraint violation is then bounded below
\begin{equation}
\begin{aligned}
&\mathbb{E}[{\mathcal{J}^i_g}(\bm{\pi})]-\bm{b}_i \leq \big[\mathcal{J}^i_g(\bm{\pi^*})-\bm{b}_i\big]+\big[\mathbb{E}[{\mathcal{J}^i_g}(\bm{\pi})]-\mathcal{J}^i_g(\bm{\pi^*})\big]\\
&\leq \bm{\tau}_{c}+\big[\mathbb{E}[{\mathcal{J}^i_g}(\bm{\pi})]-\mathcal{J}^i_g(\bm{\pi^*})\big]\\
&\leq \bm{\tau}_{c}+\Theta(\frac{1}{(1-\gamma)\sqrt{T}})+\Theta(\frac{1}{(1-\gamma)Tm^{\frac{H}{4}}}\sqrt{\log{\frac{1}{\delta}}})
\end{aligned}
\label{Global_convergence_proof6}
\end{equation}
where we have $\bm{\tau}_{c}=\Theta(\frac{1}{(1-\gamma)\sqrt{T}})+\Theta(\frac{1}{(1-\gamma)Tm^{\frac{H}{4}}})$, therefore, we obtain:
\begin{equation}
\begin{aligned}
\mathbb{E}[{\mathcal{J}^i_g}(\bm{\pi})]-\bm{b}&_i\leq\Theta(\frac{1}{(1-\gamma)\sqrt{T}})\\
&+\Theta(\frac{1}{(1-\gamma)Tm^{\frac{H}{4}}}\sqrt{\log{\frac{1}{\delta}}})
\end{aligned}
\label{Global_convergence2}
\end{equation}
where \autoref{Global_convergence2} suggests that there exists a sublinear rate of $\Theta(1/\sqrt{T})$ if the constraints are violated with an error of $\Theta(1/{m^{\frac{H}{4}}})$, with probability at least $1-\delta$. \hfill $\blacksquare$

\clearpage
\section{Experiment Supplementary}
\label{Section_App_Exp_set}
\subsection{Experimental Setting}
\label{Subsec_exp_set}
The parameter setting of AWaVO is shown in \autoref{model_parameters}, and the technical specification of the quadrotor is shown in \autoref{hardware_spec}.

\begin{table}[H]
    \centering
    \caption{Parameter Setting of AWaVO}
    \label{model_parameters}
    \begin{adjustbox}{width=0.8\textwidth}
    \begin{tabular}{c c c}
    \toprule
    \textbf{Parameters} & \textbf{Definition} & \textbf{Values} \\
    \hline
    $l_{\mu,cart}$ & Learning rate of actor in Cartpole \cite{xu2021crpo} & 0.0005 \\
    $l_{\theta,cart}$ & Learning rate of critic in Cartpole \cite{xu2021crpo} & 0.0005 \\
    $l_{\mu,acro}$ & Learning rate of actor in Acrobot \cite{xu2021crpo} & 0.005 \\
    $l_{\theta,acro}$ & Learning rate of critic in Acrobot \cite{xu2021crpo} & 0.005 \\
    $l_{\mu,guard}$ & Learning rate of actor in Walker and Drone \cite{zhao2023guard} & 0.001 \\
    $l_{\theta,guard}$ & Learning rate of critic in Walker and Drone \cite{zhao2023guard} & 0.001 \\
    $\mu$ & \makecell[c]{Actor neural network: fully connected with $H$ \\hidden layers ($m$ neurons per hidden layer)} & - \\
    $\theta$ & \makecell[c]{Critic neural network: fully connected with $H$ \\hidden layers ($m$ neurons per hidden layer)} & - \\
    $D$ & Replay memory capacity & $10^6$ \\
    $B$ & Batch size & 128 \\
    $\gamma$ & Discount rate & 0.998 \\
    $m$ & the width of neural network & 128 \\
    $H$ & the layer of neural network & 2 \\
    $T$ & Length in each episode & 500 \\
    $N$ & Time steps & 20 \\
    \bottomrule
    \end{tabular}
    \end{adjustbox}
\end{table}

\begin{table}[H]
    \centering
    \caption{Technical Specification of Hardware}
    \label{hardware_spec}
    \begin{adjustbox}{width=0.7\textwidth}
    \begin{tabular}{c c c}
    \toprule
    \textbf{No.} & \textbf{Component} & \textbf{Specific Model} \\
    \hline
    1 & Frame & QAV250 \\
    2 & Sensor - Depth Camera & Intel RealSense D435i \\
    3 & Sensor - Down-view Rangefinder & Holybro ST VL53L1X \\
    4 & Flight Controller & Pixhawk 4 \\
    5 & Motors & T-Motor F60 Pro IV 1750KV \\
    6 & Electronic Speed Controller & BLHeli-32bit 45A 3-6s \\
    7 & On-board Companion Computer & \makecell[c]{DJI Manifold 2-c\\ (CPU Model: Intel Core i7-8550U)}  \\
    8 & Mounts & \makecell[c]{3D Print for Sensors/\\Computer/Controller/Battery} \\
    \bottomrule
    \end{tabular}
    \end{adjustbox}
\end{table}

\subsection{Task Descriptions in the Simulated Platforms}
\label{Subsection_task}
\textbf{Acrobot and Cartpole tasks in OpenAI Gym.} \quad In Cartpole \cite{brockman2016openai}, the pole movement is constrained within the range of $[-2.4, 2.4]$. Each episode has a maximum length of $200$ steps and is terminated if the angle of the pole exceeds $12$ degrees. During training, the agent receives a reward of $+1$ for each step taken. However, it incurs a penalty of $+1$ if (\textit{i}) it enters the areas $[-2.4, -2.2]$, $[-1.3, -1.1]$, $[-0.1, 0.1]$, $[1.1, 1.3]$, or $[2.2, 2.4]$, or (\textit{ii}) the angle of the pole exceeds $6$ degrees.

In Acrobot \cite{brockman2016openai}, the agent is rewarded for swinging the end-effector at a height of $0.5$, where each episode has a maximum length of $500$ steps. Conversely, it faces a penalty if (\textit{i}) torque is applied to the joint when the first pendulum swings in an anticlockwise direction, or (\textit{ii}) if the second pendulum swings in an anticlockwise direction with respect to the first pendulum.

\textbf{Walker and Drone tasks in GUARD.} \quad Walker \cite{zhao2023guard}, a bipedal robot, comprises four primary components: a torso, two thighs, two legs, and two feet. Notably, unlike the knee and ankle joints, each hip joint possesses three hinges in the x, y, and z coordinates, enabling versatile turning. Maintaining a fixed torso height, Walker achieves mobility through the control of 10 joint torques.

Drone in GUARD \cite{zhao2023guard} is designed to emulate a quadrotor, simulating the interaction between the quadrotor and the air by applying four external forces to each of its propellers. These external forces are configured to counteract gravity when no control actions are applied. To maneuver in three-dimensional space, the Drone utilizes four additional control forces applied to its propellers.

\subsection{Further Details on Constraint Limit Setting}
\label{constraint_limit_setting}
In accordance with the benchmark \cite{xu2021crpo}, we established the constraint limit as $50$ in Acrobot, as depicted in Figure~\autoref{learning_cur_Acrobot}. In the remaining scenarios, namely Cartpole in Figure~\autoref{learning_cur_cartpole}, Walker in Figure~\autoref{learning_cur_Walker}, Drone in Figure~\autoref{learning_cur_drone}, and the real quadrotor in \autoref{real_flight_tasks_conv}, the constraint limit serves as a lower boundary, indicating the level of tolerance the constraints can endure. The agent's stable performance for specific tasks occurs when it operates below this constraint limit. We hypothesize that there may be potential benefits in establishing a fixed limit, $b_i$, by decoupling the cumulative value into specific fixed limits. This is left for future work.

\subsection{Further Discussion on Probabilistic Interpretation of Sequential Decisions}
\label{discuss_probabilistic_interpretation}
\textbf{Curves in \autoref{Pro_interpret} (a) and (b).} \quad The curves in \autoref{Pro_interpret} (a), provided as a reference, give the estimated values of external aerodynamic forces (winds) in real-time. These estimates are derived from the signals collected from onboard sensors. In \autoref{Pro_interpret} (b), the curves illustrate the quantitative impact of external forces $L_0$ on current sequential decisions, specifically, the planned trajectory $\tau$. This impact is quantified as parts of pulse width modulation signals that are fed into the motors to either resist or cooperate with the measured (or identified) aerodynamic forces. \autoref{Pro_interpret} (b) aims to quantitatively interpret and visually convey the decision-making process in response to external forces.

\autoref{Pro_interpret} (b) becomes particularly important when the agent makes sub-optimal decisions leading to events like quadrotor crashes or collisions. These curves prove valuable for interpreting and performing quantitative analyses of distinct environmental factors, such as winds and obstacles, allowing for an understanding of their magnitudes of influence on the current decision-making process.

\textbf{An instance to interpret \autoref{Pro_interpret} (b).}  \quad In the case of Reference State 02 (RS 02), located in an area with a combination of wind and obstacles, both aerodynamic effects (i.e., external forces) from winds and obstacles act simultaneously on the quadrotor. In Flight Task 3 (FT 3), represented by the red curve, we can observe the influence of external forces (i.e., aerodynamic effects from winds and obstacles) on the current trajectory planning decisions. The value is approximately $0.40$ at RS 02, implying that the ongoing trajectory planning decisions have a $\sim40\%$ probability of being influenced by the aerodynamic effects. 

Comparing FT 1 and FT 2, where the values at RS 02 are approximately $0.20$ (FT 1) and $0.18$ (FT 2), respectively, we can decouple the aerodynamic effects generated by the wind on the body (FT 1) and obstacles (FT 2). Quantitatively, at RS 02, situated in an area with a mix of wind and obstacles, the red $p(\tau|L_0)_{FT3}$ is approximately equal to the sum of $p(\tau|L_0)_{FT1}$ (only wind) and $p(\tau|L_0)_{FT2}$ (only obstacles).

\subsection{Scalability}
\label{scalability}
To explore the full scalability of our methodology, we have implemented it on the Lorenz 96 system \cite{lorenz1998optimal,gorbach2017scalable}, characterized by the following equations: ${dx_i}/{dt} = (x_{i+1} - x_{i-2}) \cdot x_{i-1} - x_i + f_\theta$, where $f_\theta$ represents a scalar forcing parameter, and $x_{-1}=x_{I-1},x_0=x_I,x_{I+1}=x_1$, with $I$ denoting the number of states in the deterministic system. This system serves as a simplified model for weather forecasting. We choose it as a versatile framework for expanding the number of states in our inference task. In our experiment, we range from 50 to 1000 states, with approximately one-third of the states randomly designated as unobserved.

\begin{figure}[h]
    \centering
    \includegraphics[scale=0.55]{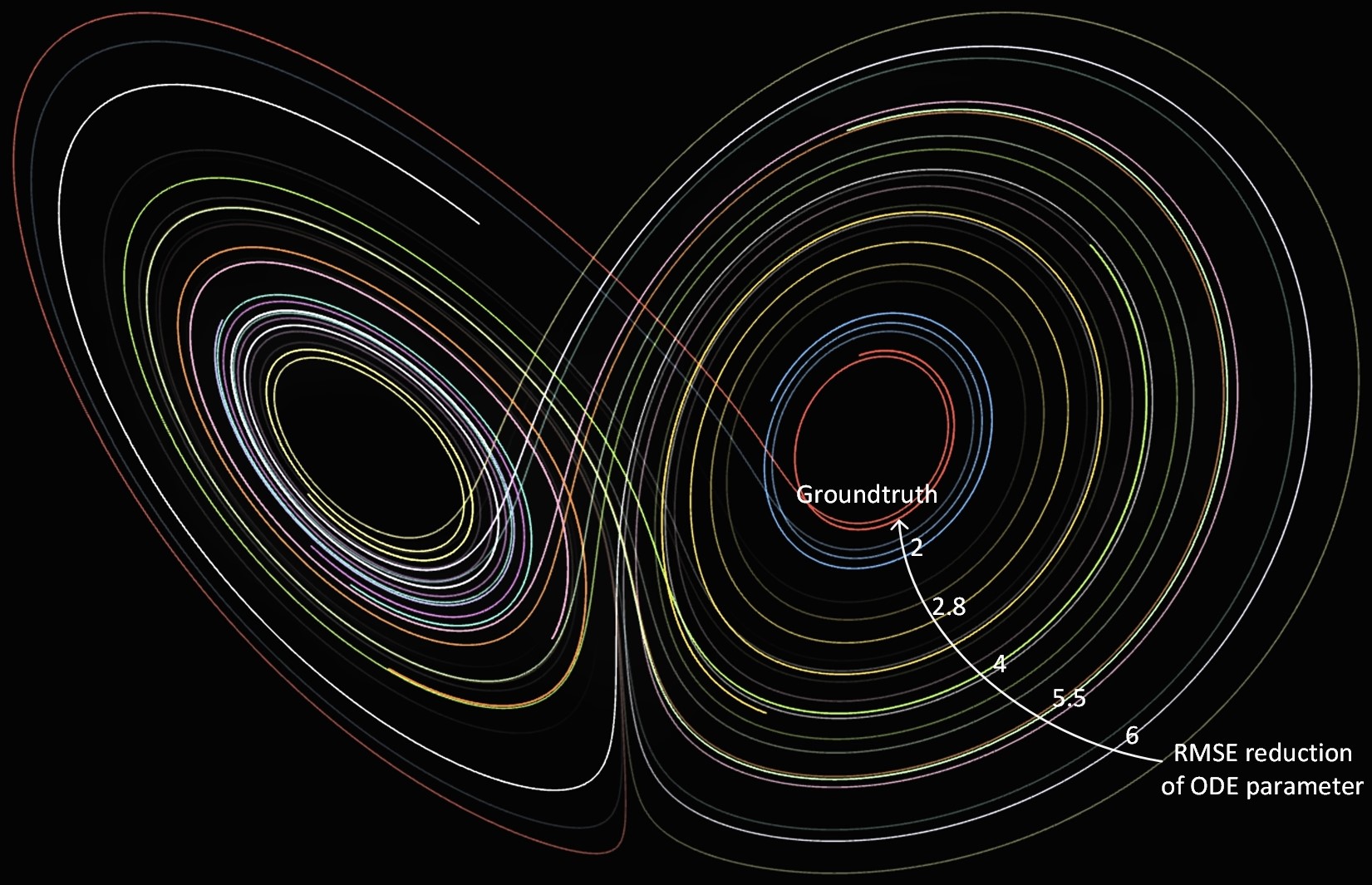}
    \caption{Scalability performance on Lorenz 96 system \cite{lorenz1998optimal,gorbach2017scalable}.}
    \label{lorenz_performance}
\end{figure}

\begin{figure}[h]
    \centering
    \includegraphics[scale=0.2]{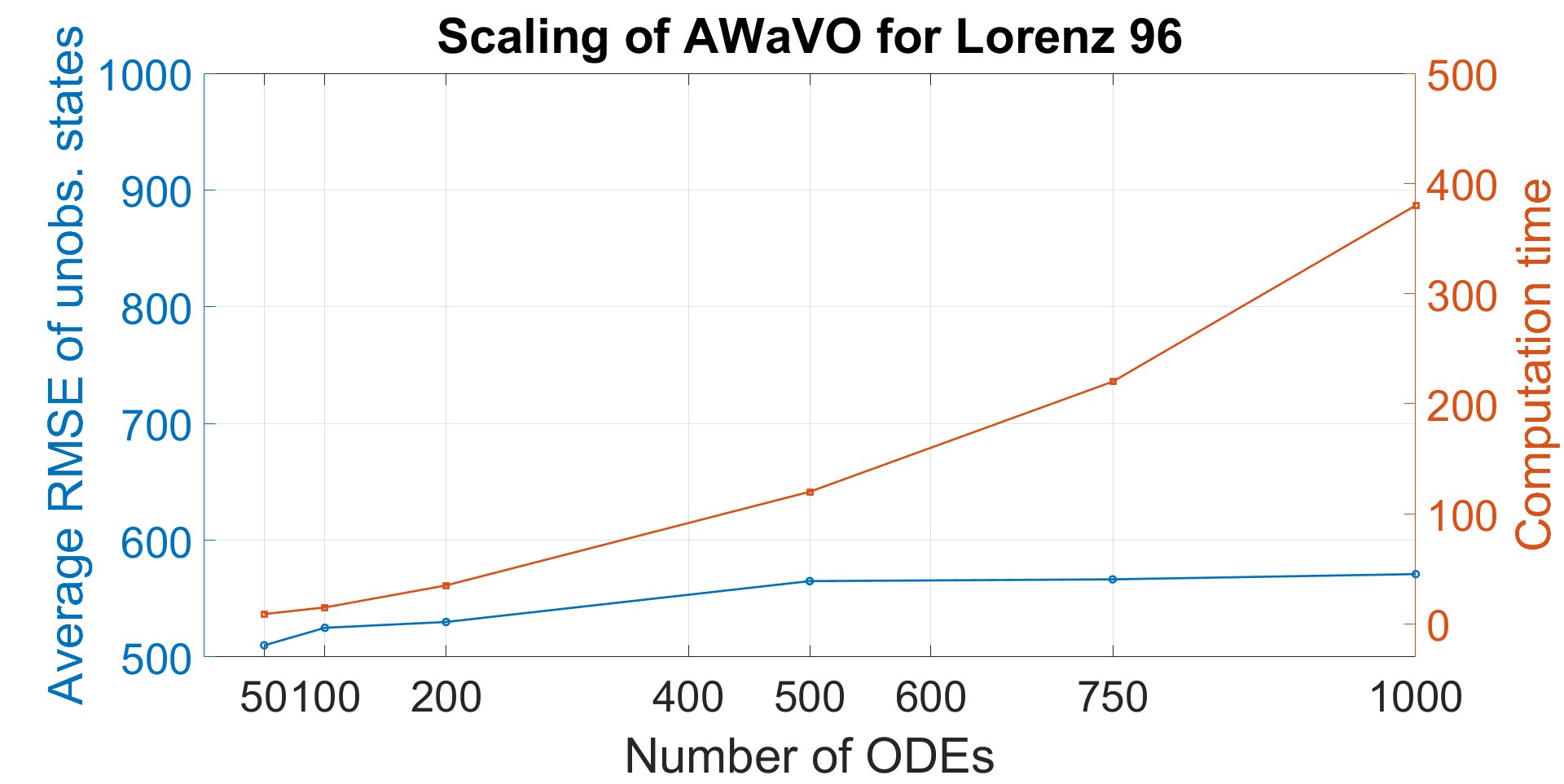}
    \caption{Average RMSE and computation time of the unobserved state for scalability.}
    \label{lorenz_statistic}
\end{figure}

We show the results in \autoref{lorenz_performance} and \autoref{lorenz_statistic}. Our methodology successfully infers a system with 1000 states in less than 400 seconds (shown in \autoref{lorenz_statistic}). From visual inspection, we observe that the unobserved states are inferred, with the approximation error remaining consistent regardless of the problem's dimensionality. Given that many real-world RL problems involve state spaces significantly larger than 1,000 states, our future work will explore such scalability concerns on larger models.

\end{document}